\documentclass{article}
\usepackage{iclr2027_conference,times}
\usepackage[T1]{fontenc}

\usepackage{amsmath,amssymb}

\usepackage{amsmath,amsfonts,bm}

\def\eqref#1{equation~\ref{#1}}

\def\1{\bm{1}}

\DeclareMathAlphabet{\mathsfit}{\encodingdefault}{\sfdefault}{m}{sl}
\SetMathAlphabet{\mathsfit}{bold}{\encodingdefault}{\sfdefault}{bx}{n}

\usepackage{hyperref}
\usepackage{url}
\usepackage{graphicx}
\usepackage{booktabs}
\usepackage{array}
\makeatletter
\newenvironment{tableH}{\par\addvspace{\intextsep}\noindent\begin{minipage}{\linewidth}\def\@captype{table}}{\end{minipage}\par\addvspace{\intextsep}}
\newenvironment{figureH}{\par\addvspace{\intextsep}\noindent\begin{minipage}{\linewidth}\def\@captype{figure}}{\end{minipage}\par\addvspace{\intextsep}}
\makeatother
\usepackage{multirow}
\usepackage{xcolor}
\usepackage[capitalize,noabbrev]{cleveref}
\usepackage{tikz}
\usetikzlibrary{arrows.meta,positioning,shapes.geometric,fit,backgrounds,calc}

\definecolor{sgink}{HTML}{3D3A38}
\definecolor{sgtext}{HTML}{7C7671}
\definecolor{sgline}{HTML}{C9C2BC}
\definecolor{sgcream}{HTML}{FAF7F3}
\definecolor{sgmint}{HTML}{B8DFD0}
\definecolor{sgmintink}{HTML}{3F7A66}
\definecolor{sgmintband}{HTML}{F1F8F5}
\definecolor{sgrose}{HTML}{F3BCC6}
\definecolor{sgroseink}{HTML}{B05568}
\definecolor{sgroseband}{HTML}{FDF3F5}
\definecolor{sglilac}{HTML}{C9BCE6}
\definecolor{sglilacink}{HTML}{6A57A3}

\newcommand{\sg}{\textsc{SimuGuard}}

\newcommand{\todo}[1]{}
\newcommand{\pending}[1]{}

\newcommand{\hid}[1]{}

\title{When the Simulator Fails, Not the Policy:\\Auditing Contact Artifacts\\in Simulated Robot Benchmarks}

\author{Alex Zhou\quad Junhao Wu\quad Xueyang Zhou\quad Yixiao Huo\quad Hongcheng Gao\\
\textbf{Di Zhu\quad Zhenghao Zhang\quad Yongchao Chen$^{\dagger}$}\\[4pt]
College of AI, Tsinghua University\\
$^{\dagger}$Corresponding author: \texttt{chenyongchao@mail.tsinghua.edu.cn}}

\iclrfinalcopy
\begin{document}

\maketitle
\lhead{Preprint}

\begin{abstract}
Simulation benchmarks score robot policies by success rate, and every failed episode counts against the policy. Testing over ten vision language action and world action models on five benchmarks, we find that failures in which the policy has placed the object and the simulator then throws it out are frequent, and that the benchmark's own scripted demonstrations fail the same way. We present \sg{}, an auditing plugin for LIBERO, RoboTwin, RoboCasa, RoboDojo and RoboLab. At every physics step, it records the state and flags an object that starts moving at a contact faster than the robot's contact or gravity account for. It replays a recorded episode bit for bit, and it reruns the episode with one solver parameter changed. With it we trace each such episode to a false contact. A fixed threshold in PhysX's collision detection makes the engine report an overlap of about two centimeters between the placed object and the thin wall of its container, where the two do not touch. The solver removes the whole overlap in one step and throws the object out at several meters per second. With a smaller threshold, LingBot-VA succeeds in 98 of 100 new episodes against 89, and no episode has an artifact. Across five models, the ranking in the repaired engine follows the ranking on a real robot. \sg{} also finds two other simulation errors: objects that start inside their support, and objects stuck to an open gripper. \sg{} lets a benchmark separate what the policy did from what the solver did.
\end{abstract}
\section{Introduction}
\label{sec:intro}

Simulation benchmarks give a repeatable way to compare robot policies, from vision language action (VLA) models to world action models~\citep{brohan2023rt2,brohan2023rt1,octo2024,kim2024openvla,black2024pi0,bjorck2025gr00t,geminirobotics2025,li2026lingbotva}. LIBERO~\citep{liu2023libero}, RoboCasa~\citep{nasiriany2024robocasa} and RoboTwin~\citep{mu2025robotwin,chen2025robotwin2} fix the tasks and the protocol and treat the simulator as faithful physics. Their success rates record which episodes failed, not why. A failure caused by the simulator's treatment of contact enters the same count as a failure caused by the policy. Testing over ten world action models and VLAs in simulation, we find that such failures are frequent: the policy places the object, and the simulator throws it out. Such ejections, often called physics explosions, are a known effect of rigid body solvers. We measure how often they decide benchmark episodes and trace them to a false contact from a fixed threshold in PhysX's collision detection.

\begin{figure}[t]
  \centering
  \includegraphics[width=\linewidth]{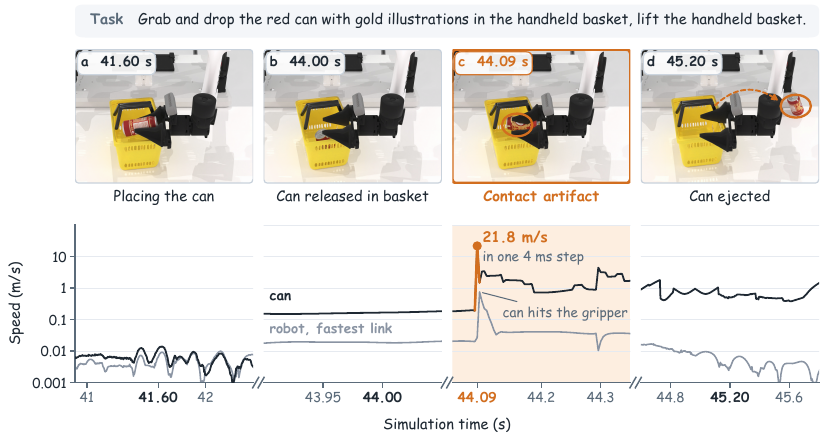}
  \caption{\textbf{A contact artifact turns a correct placement into a failure.} LingBot-VA places the can in the basket (a, b). Within one physics step, the can's speed rises from 0.2 to 22\,m/s while no robot link touches it (c). The can flies out and the benchmark records a failure (d). The time axis has four linear segments, one under each frame.}
  \label{fig:ejection}
\end{figure}

\begin{figure}[t]
  \centering
  \includegraphics[width=\linewidth]{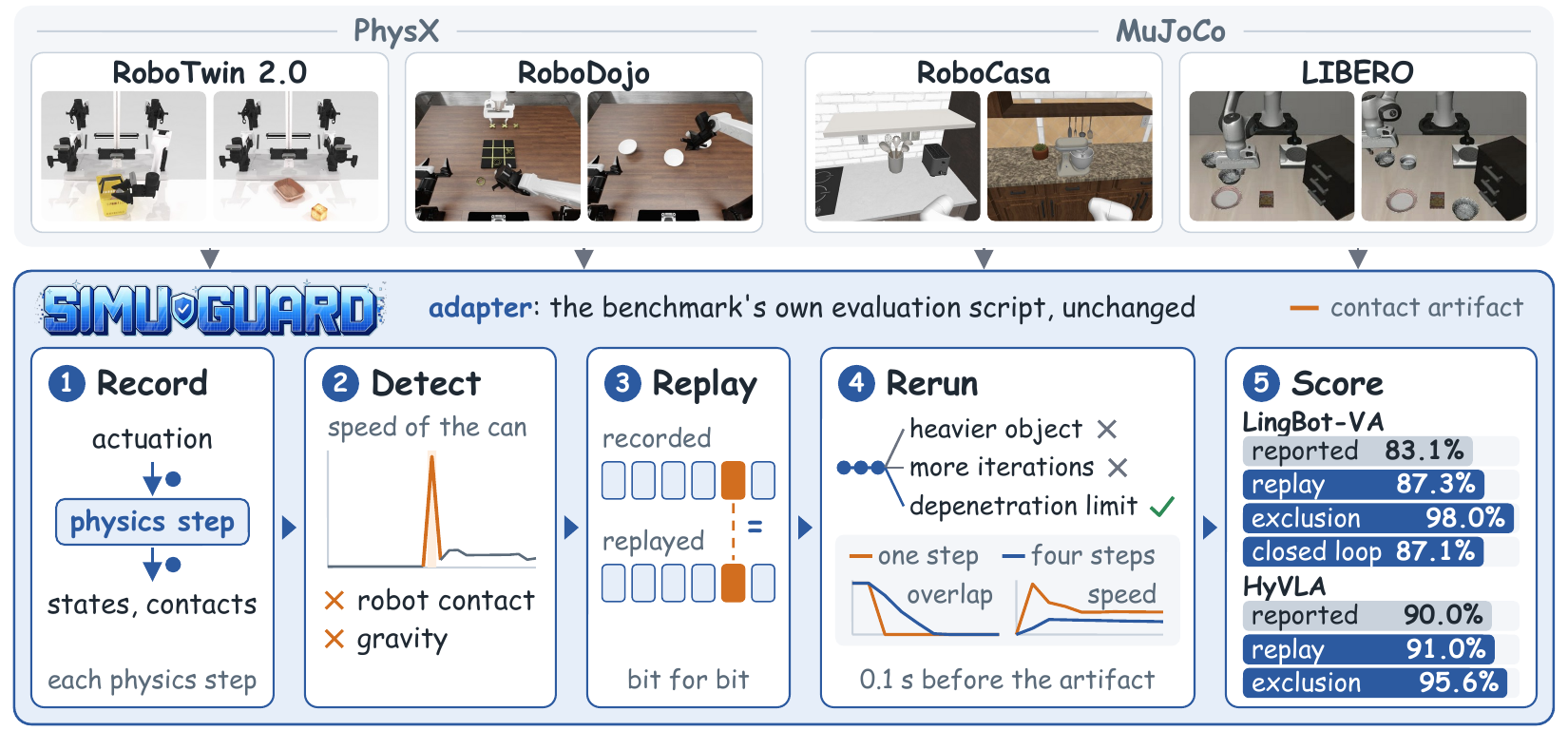}
  \caption{\textbf{\sg{} audits a simulated evaluation.} \sg{} records every physics step, detects contact artifacts, replays the episode exactly, and reruns it with one simulation parameter changed. Inset: PhysX removes a 20\,mm overlap in one step, or over four steps with the depenetration limit. Score: the four ways of scoring the five container tasks (\cref{sec:method-score}).}
  \label{fig:teaser}
\end{figure}

\begin{figure}[t]
  \centering
  \includegraphics[width=\linewidth]{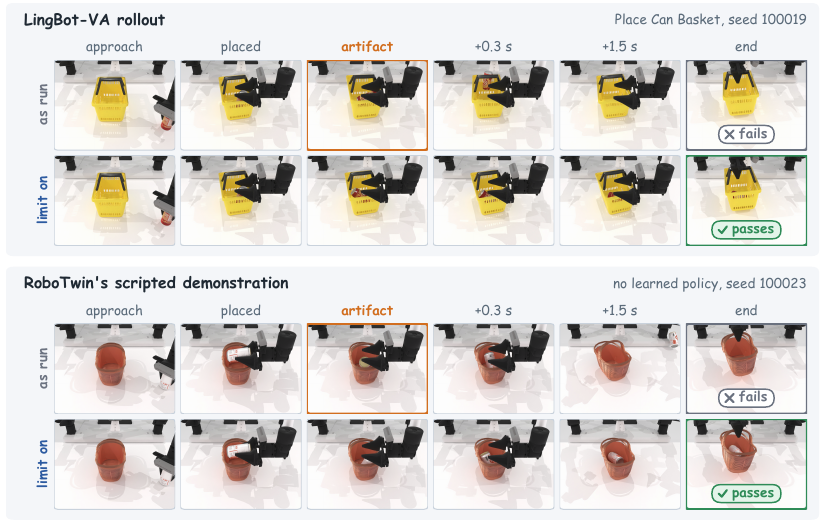}
  \caption{\textbf{Two episodes whose outcome the replay flips.} Each is shown as run and as replayed with the depenetration limit. \emph{Top:} LingBot-VA on Place Can Basket. \emph{Bottom:} one of RoboTwin's own scripted demonstrations.}
  \label{fig:trajectories}
\end{figure}

In an episode of RoboTwin~2.0's Place Can Basket, LingBot-VA~\citep{li2026lingbotva} places a can inside the basket and releases it. Within a single physics step, PhysX reports the can inside the basket wall, where the step before it reported no overlap. The can's speed jumps from 0.2 to 22\,m/s while no robot link touches it, the can flies out, and the benchmark records a failure (\cref{fig:ejection}). Replayed with the same recorded actions and the overlap removed over several physics steps instead of one, the episode passes the benchmark's own success check (\cref{fig:trajectories}). We call a contact response that sets an object moving faster than the robot's contact or gravity account for a contact artifact. It lasts a single physics step, between two camera frames, so a video shows the can flying but not what launched it, and the success check cannot tell whether the policy or the solver decided the episode.

\sg{} is an auditing plugin (\cref{fig:teaser}). An adapter attaches it to the benchmark's own evaluation script; we provide adapters for LIBERO, RoboTwin, RoboCasa, RoboDojo and RoboLab~\citep{yang2026robolab}. At every physics step it records the actuation, the object states and the contacts, and flags an object that starts moving at a contact faster than the robot's contact or gravity account for. It replays a recorded episode bit for bit by rebuilding the seeded scene and applying the recorded actuation from the first step. A rerun is a replay with one solver parameter changed, with the recorded actuation and the success check fixed. Prior work adds metrics to the success rate~\citep{wang2025roboeval,snyder2026binary,jiang2026benchmarking} or measures how solvers depart from physics~\citep{erez2015simulation,lelidec2023contact}; neither connects an anomalous contact in a rollout to the outcome of that rollout.

Our contributions are:
\begin{itemize}
  \item \textbf{A finding.} In simulated benchmarks, the policy often places the object and the contact solver then ejects it, and the benchmark scores a failure. The benchmark's own scripted demonstrations fail the same way. We trace the ejection to a fixed threshold in PhysX's collision detection; with a smaller threshold, the artifacts disappear. \sg{} also finds two other simulation errors: objects that start inside their support on LIBERO, and objects stuck to an open gripper on RoboLab.
  \item \textbf{A tool.} \sg{} monitors every physics step, replays a recorded episode bit for bit, and reruns it with one solver parameter changed.
  \item \textbf{An evaluation method.} We correct the score by replaying each flagged episode with the depenetration velocity of the ejected body limited around the artifact, and compare it with three other ways of scoring.
\end{itemize}

\section{Related Work}
\label{sec:related}

\paragraph{Simulated evaluation of robot policies.}
VLA policies adapt large pretrained backbones to produce robot actions~\citep{brohan2023rt2,brohan2023rt1,oxe2024,octo2024,kim2024openvla,black2024pi0,bjorck2025gr00t}, and world action models predict future video together with the actions~\citep{du2023unipi,wu2024gr1,cheang2024gr2,hu2025vpp,li2025uva,li2026lingbotva}. Both are compared on simulated suites built on two families of physics engines: LIBERO~\citep{liu2023libero} and RoboCasa~\citep{nasiriany2024robocasa} run on MuJoCo~\citep{todorov2012mujoco}, while ManiSkill~\citep{mu2021maniskill1,gu2023maniskill2,tao2024maniskill3} and RoboTwin~\citep{mu2025robotwin,chen2025robotwin2} run on SAPIEN~\citep{xiang2020sapien}, which uses PhysX, as does Isaac Lab~\citep{mittal2023orbit}. SimplerEnv~\citep{li2024simplerenv} tests whether their rankings carry over to real robots, and \citet{jiang2026benchmarking} audit five of these suites and find shortcut solutions, statistically insignificant gains and overfitting to the test distribution; they ask what a score says about the policy and take the simulator as given. SafeVLA-Bench~\citep{fan2026safevla} instruments existing benchmarks after the fact to report successes that violate a safety specification, and also takes the simulated physics as correct. All of these suites score each episode with a scripted check on the final state and do not examine the physics that produced it.

\paragraph{Contact simulation.}
Rigid body engines approximate contact differently. MuJoCo treats it as a soft constraint that removes an overlap over a time constant of several physics steps~\citep{todorov2012mujoco}; PhysX treats it as a hard constraint and pushes overlapping bodies apart within the step, up to a configurable maximum speed, and its documentation records that bodies created in overlap may be separated violently and names the depenetration limit as the remedy~\citep{physx54}. Compliant and pressure field models~\citep{castro2023convex,elandt2019hydroelastic} make other choices. The violent separation is a case of numerical instability in physics simulation, studied for cloth~\citep{baraff1998large} and fluids~\citep{stam1999stable}. For rigid bodies, staggered projections resolve frictional contact accurately~\citep{kaufman2008staggered}, and Rigid IPC guarantees that no step ends in an intersection~\citep{ferguson2021rigidipc}. The benchmarks we study run on PhysX and MuJoCo, which give no such guarantee. Engines disagree on the same scenes~\citep{erez2015simulation} and on recorded real impacts~\citep{acosta2022validating}, contact models trade physical consistency for tractability~\citep{lelidec2023contact}, and convex decomposition of assets adds its own approximation~\citep{wei2022coacd}. These studies compare engines in isolated scenes or document a known behavior; none follows one contact response inside a fixed benchmark to the outcome of the episode, which is what \sg{} does (\cref{app:related}).

\paragraph{Reliability of benchmark numbers.}
Several studies ask what a success rate measures and how many trials it needs~\citep{kressgazit2024evaluation,vincent2024generalizable}. RoboEval~\citep{wang2025roboeval} adds behavioral metrics because equal success rates can hide different executions; \citet{snyder2026binary} replace binary outcomes with graded progress; LIBERO-PRO~\citep{zhou2025liberopro} and LIBERO-Plus~\citep{fei2025liberoplus} show that high scores often do not survive changes to objects, layouts and observations; and \citet{henderson2018matters}, \citet{agarwal2021precipice} and \citet{bouthillier2021variance} show how variance and point estimates distort comparisons. These works improve how outcomes are aggregated and take the outcome of each episode as given. We test the outcome itself: \sg{} traces one contact response to the outcome of the episode that contains it, with a replay anyone can verify, and compares four ways of scoring a benchmark once its artifacts are known.

\section{\sg{}}
\label{sec:method}

\sg{} has three requirements: it must not change what it measures, it must reproduce an episode exactly so that a difference after an intervention belongs to the intervention, and it must run with the benchmark's own evaluator so that its findings apply to the reported numbers.

\subsection{Setting}
\label{sec:method-setting}

A benchmark evaluates a policy on a sequence of seeds. Each seed fixes the scene, and the evaluator runs the policy's actions through the simulator until a scripted success check passes or the action budget is spent. The simulator advances in physics steps of a few milliseconds (4\,ms in RoboTwin, 2\,ms in LIBERO and RoboCasa); one policy action spans tens of them. Of $N$ scored episodes, $S$ succeed, and the benchmark reports $S/N$.

A contact artifact is motion that starts at a contact and that neither the robot's contact nor gravity accounts for; in this paper the term names what the detector of \cref{sec:method-detect} flags, and \cref{sec:exp-reliability} measures its precision against human judgment. Let $A$ be the set of scored episodes that contain one. \Cref{fig:teaser} shows the five steps that follow.

\subsection{Recording without intrusion}
\label{sec:method-record}

\sg{} wraps the physics step of the running scene, so the evaluator, the benchmark and the policy run unmodified in their own process. A thin adapter per engine provides the wrapper, around the step function of SAPIEN (RoboTwin), Isaac Sim (RoboDojo, RoboLab) or robosuite, which runs LIBERO and RoboCasa on MuJoCo. Before each step, \sg{} records the actuation: joint drive targets and forces in PhysX; control inputs, applied forces and mocap poses in MuJoCo. It also records any state that benchmark code writes between two steps, which it must replay (\cref{sec:method-replay}). After each step, it records the pose and velocity of every body and the contacts, merged per pair of bodies, with impulse and penetration depth. A task specification names the objects a task manipulates and the containers they go into. The evaluator, the benchmark and the policy code are untouched, and with the monitor on or off, all 59 tracked object trajectories in our check are identical; the monitor adds 7.5 to 11.6\,ms to each 4\,ms step in SAPIEN (\cref{app:impl}).

\subsection{Detecting contact artifacts}
\label{sec:method-detect}

Detection has three stages, with the same thresholds for every task (\cref{app:detector}). The first finds motion that starts at a contact. An object becomes a candidate when it touched another body within the last 40\,ms, moves faster than 0.5\,m/s, and changes velocity by at least 0.35\,m/s within one 4\,ms step (the thresholds on velocity change scale with the step). The candidate is confirmed if within 240\,ms it travels at least 4\,cm and either flies free for 16\,ms or stays in violent contact with its container.

The second stage removes motion that gravity explains. An object dropped into a bin also moves fast right after a contact. For each confirmed event, \sg{} compares the object's peak speed in the 0.1\,s after onset with the free fall speed $\sqrt{2g\,\Delta z}$ for the height it dropped before that peak, and keeps the event only if the ratio exceeds 1.5.

The third stage removes motion that the robot transmits through another body. A basket that the robot's base shoves across the floor carries the bread inside it at the basket's speed. \sg{} finds every body that touched both the object and a robot link in the 40\,ms before onset, and discards the event if the object's peak speed is less than 1.5 times the fastest such carrier's speed before onset. An event that passes all three stages is a candidate contact artifact. The stages compare speeds, not the work done on the object, so human review is the final standard: raters judged the flagged clips frame by frame (\cref{sec:exp-reliability}).

\subsection{Exact replay}
\label{sec:method-replay}

\sg{} replays an episode by rebuilding the scene from its seed and applying the recorded actuation from the first physics step. Because the whole trajectory is regenerated from its initial conditions, the replay matches the recording bit for bit (\cref{sec:exp-reliability}). Three details make this hold. PhysX cannot restore a saved state exactly once bodies touch, since trajectories restored through its state interface diverge by millimeters to centimeters, so \sg{} always replays from the start. Benchmark code changes the state between physics steps: a RoboCasa fixture sets its door joint, and robosuite~1.4, which LIBERO uses, recomputes the dynamics before each step and so rewrites the solver's warm start; \sg{} records both and writes them back. Finally, outcomes depend on values outside the physics: RoboTwin's success checks read the gripper opening the robot was commanded to, which \sg{} restores at each step from the recorded drive targets.

\subsection{Controlled reruns}
\label{sec:method-rerun}

A rerun is a replay with one simulation parameter changed: the mass of the manipulated object, the number of solver iterations, the maximum speed at which PhysX separates overlapping bodies (its maximum depenetration velocity), or the time constant of MuJoCo's contacts. A rerun either applies the change from the first physics step, or switches it on 0.1\,s before the episode's first contact artifact. In the second form, everything before the switch is identical to the recording, which \sg{} verifies bit for bit. In its long version the limit acts on every body until the episode ends. In its short version, which we use for the scores, it acts only on the ejected body, until 0.2\,s after the artifact. The benchmark's own success check scores every rerun, and the evaluator stops at the first success, as in the original evaluation. A third form keeps the change on for a new closed loop rollout on the same seed; the policy samples new actions, so the episode is no longer comparable one by one with the recording.

\subsection{Scoring}
\label{sec:method-score}

Once the flagged episodes $A$ of an evaluation are known, and if only they can change, the success rate under a physics that never ejects lies between $(S - S_A)/N$, every flagged episode failing, and $(S - S_A + |A|)/N$, every flagged episode succeeding, where $S_A$ counts the successes in $A$; the one assumption, that no unflagged episode changes, fails where the detector misses an artifact. On the five RoboTwin tasks of \cref{sec:exp-where} it is $[75.8\%, 98.5\%]$ for LingBot-VA and $[86.0\%, 96.0\%]$ for HyVLA. Each of the four rates we report is a point inside it under one further assumption (\cref{fig:teaser}, Score). The \emph{reported} rate $S/N$ assumes that every ejection is the policy's fault. The \emph{replay} rate replaces the recorded outcome $y_e \in \{0,1\}$ of each flagged episode by the outcome $r_e$ of its recorded actions under the short limit,
\begin{equation}
  \hat{s} \;=\; \frac{1}{N} \Big( S \;+\; \sum_{e \in A} (r_e - y_e) \Big),
  \label{eq:corrected}
\end{equation}
and assumes that those actions remain the policy's choice once the ejection is removed. The \emph{exclusion} rate $(S - S_A)/(N - |A|)$ assumes that the flagged episodes are as hard as the others. The \emph{closed loop} rate is the success rate of new rollouts with the limit kept on, and assumes that the limit changes nothing the task needs. We report the interval and the artifact count, and use the replay rate as the point estimate.

\section{Experiments}
\label{sec:experiments}

\paragraph{Setup.}
We run RoboTwin~2.0~\citep{chen2025robotwin2} through its official evaluation script in its clean configuration (no background, table or lighting randomization, seen instructions), on SAPIEN~\citep{xiang2020sapien} with PhysX and the benchmark's 4\,ms physics step. We audit two kinds of policy, each with its RoboTwin checkpoint: LingBot-VA~\citep{li2026lingbotva}, a world action model that predicts future video together with its actions, on all 50 tasks, and HyVLA~\citep{zhang2026hyvla}, a vision language action model, on the five container tasks of \cref{sec:exp-where} and six other tasks. Before the policy runs on a seed, RoboTwin runs a scripted demonstration and scores the policy only on seeds where the demonstration passes; every rate in this paper is over scored episodes, and skipped seeds do not enter $N$. On MuJoCo, with a 2\,ms physics step, we run $\pi_{0.5}$~\citep{intelligence2025pi05} on the 40 tasks of LIBERO and XR-1~\citep{fan2025xr1} on all 50 tasks of its RoboCasa evaluation set. RoboDojo~\citep{chen2026robodojo} runs on Isaac Sim with PhysX and a 4\,ms physics step; we run XR-1 on four of its tasks, and eight models with real robot results on seven tasks. \sg{} monitors every episode, and all analysis uses the stored recordings. Episode counts, versions and checkpoints are listed in \cref{app:runs}.

\subsection{Where contact artifacts occur}
\label{sec:exp-where}

\begin{table}[t]
  \centering
  \caption{\textbf{Success rates on RoboTwin~2.0} in percent, under the four scores of \cref{sec:method-score}. \emph{Rep.}: reported. \emph{Replay}: flagged episodes replayed with the short limit. \emph{Excl.}: flagged episodes dropped. \emph{CL}: new closed loop rollouts with the limit on. Counts: \cref{tab:counts}.}
  \label{tab:scores}
  \small
  \begin{tabular*}{\linewidth}{@{\extracolsep{\fill}}lcccccccc@{}}
    \toprule
    & \multicolumn{4}{c}{LingBot-VA (world action model)} & \multicolumn{4}{c}{HyVLA (vision language action)} \\
    \cmidrule(lr){2-5}\cmidrule(l){6-9}
    Task & Rep. & Replay & Excl. & CL & Rep. & Replay & Excl. & CL \\
    \midrule
    Place Can Basket    & 74.4 & 81.1  & 96.2  & 90.0  & 80.0  & 75.0  & 81.3  & 80.0 \\
    Stack Bowls Three   & 76.0 & 80.0  & 100.0 & 100.0 & 90.0  & 85.0  & 100.0 & 100.0 \\
    Place Object Basket & 87.5 & 87.5  & 94.1  & 50.0  & 95.0  & 95.0  & 95.0  & 45.5 \\
    Place Bread Basket  & 97.5 & 97.5  & 100.0 & 85.0  & 100.0 & 100.0 & 100.0 & 70.0 \\
    Stack Bowls Two     & 92.5 & 100.0 & 100.0 & 100.0 & 85.0  & 100.0 & 100.0 & 100.0 \\
    \midrule
    Five tasks          & 83.1 & 87.3  & 98.0  & 87.1  & 90.0  & 91.0  & 95.6  & 78.4 \\
    \bottomrule
  \end{tabular*}
\end{table}

Failures with contact artifacts concentrate in five tasks, each of which places an object into a container with thin walls (\cref{tab:scores}). Of LingBot-VA's 44 failures on these tasks, 40 contain an artifact; in its other 45 tasks, 12 of 146 failures do, on eight tasks (\cref{tab:manifest}). HyVLA shows the same pattern: 6 of its 10 failures on the five tasks contain an artifact, and none of its 38 failures on the six other tasks.

\begin{table}[t]
  \centering
  \caption{\textbf{The benchmark's own expert fails the same way.} Place Can Basket: failures with an ejection, and failures that pass when replayed with the depenetration limit.}
  \label{tab:expert}
  \small
  \begin{tabular}{@{}lcccc@{}}
    \toprule
    Actor & Episodes & Failures & With an ejection & Pass with the limit \\
    \midrule
    LingBot-VA (world action model) & 90 & 23 & 21 & 18 \\
    HyVLA (vision language action)  & 20 & 4  & 1  & 1 \\
    Scripted expert (ground truth poses) & 22  & 22 & 16 & 13 \\
    \bottomrule
  \end{tabular}
\end{table}

An artifact does not always end in failure: 19 successful LingBot-VA episodes of the five tasks contain one, 16 of them on Place Can Basket, where an ejected can may land back in the basket. Artifacts also act before the policy is scored: on Place Can Basket, 16 of the 22 seeds RoboTwin skipped because its own scripted demonstration failed contain one, and 13 pass with the limit (\cref{tab:expert}, \cref{fig:trajectories}).

\subsection{Cause: a false contact, removed within one physics step}
\label{sec:exp-cause}

\begin{table}[t]
  \centering
  \caption{\textbf{Stress test.} RoboTwin's can at 300 resting poses inside its basket, in PhysX and MuJoCo (\cref{app:stress}). \emph{Thrown}: at least 1\,m/s and 4\,cm away. \emph{Removal}: initial overlap over exit speed. RoboTwin's values in parentheses.}
  \label{tab:cause}
  \small
  \begin{tabular*}{\linewidth}{@{\extracolsep{\fill}}lcccc@{}}
    \toprule
    Engine and setting & Thrown (\%) & Exit speed (m/s) & Distance (cm) & Removal (ms) \\
    \midrule
    PhysX, RoboTwin settings           & 91 & 8.77  & 155 & 2.6 \\
    PhysX, can mass 100\,g (10\,g)     & 90 & 6.86  & 73  & 3.3 \\
    PhysX, solver iterations 32/8 (10/1) & 90 & 16.78 & 200 & 1.4 \\
    PhysX, depenetration $\le$ 1\,m/s  & 65 & 1.36  & 10  & 16.5 \\
    PhysX, depenetration $\le$ 0.1\,m/s & 0 & 0.14  & 3.7 & 169 \\
    MuJoCo, default (time constant 20\,ms) & 5 & 0.56 & 2.8 & 38.6 \\
    MuJoCo, time constant 8\,ms        & 42 & 1.99  & 3.8 & 10.6 \\
    \bottomrule
  \end{tabular*}
\end{table}

\begin{table}[t]
  \centering
  \caption{\textbf{Replays with one change,} switched on 0.1\,s before the first artifact, on LingBot-VA episodes with artifacts. \emph{Rescued}: failures that pass. \emph{Peak}: median object speed after the switch.}
  \label{tab:cause-replay}
  \small
  \begin{tabular*}{\linewidth}{@{\extracolsep{\fill}}lcccc@{}}
    \toprule
    & \multicolumn{2}{c}{Place Can Basket} & \multicolumn{2}{c}{Stack Bowls Three} \\
    \cmidrule(lr){2-3}\cmidrule(l){4-5}
    Change & Rescued & Peak (m/s) & Rescued & Peak (m/s) \\
    \midrule
    Depenetration $\le$ 1\,m/s      & 18/21 & 1.39  & 7/12 & 0.65 \\
    Object mass 100\,g              & 14/21 & 3.72  & 5/12 & 2.12 \\
    Solver iterations 32/8          & 12/21 & 11.20 & 4/12 & 7.00 \\
    \bottomrule
  \end{tabular*}
\end{table}

At the onset of an artifact, PhysX's contact report places the object deep inside the collision geometry of its container. Over the 99 artifacts on Place Can Basket, the reported overlap between can and basket at the onset step is 21.5\,mm in the median (interquartile range 11.6 to 27.9\,mm); in the 35 episodes we traced, the two convex pieces that report it are a median 15\,mm apart when the step begins. The can moves under 1\,mm in that step, so the overlap is not motion but a false contact. Its cause is a fixed threshold. PhysX~5.3.1, which RoboTwin uses, treats a triangle in its distance computation, GJK~\citep{gilbert1988gjk}, as degenerate when $|ab \times ac|^2 \le 2^{-23}$ in meters, and small pieces of a thin wall fall below it. Fed the recorded poses outside the simulator, PhysX returns all 20 false contacts we traced. With the threshold of PhysX~5.4.0, $2^{-46}$, it returns none, and 4 real contacts keep their depth (\cref{app:cause}). In the simulator, the same change cuts the scripted expert's episodes with an artifact from 18 to 1 on Place Can Basket and from 6 to 0 on Stack Bowls Three. PhysX removes the false overlap within the step: the can's velocity changes by 4.36\,m/s, close to the 5.4\,m/s at which 21.5\,mm is removed in one step.

A stress test isolates how each engine removes such an overlap (\cref{tab:cause}). We place RoboTwin's can in its basket without a robot, at 300 poses near the one recorded before the ejection of \cref{fig:ejection} (\cref{app:stress}). With RoboTwin's settings, the overlap divided by the exit speed puts PhysX's removal at 2.6\,ms, within one step, and throws the can in 91\% of poses, a median 1.55\,m away. Of the PhysX settings we tested, mass and solver iterations leave the rate at 90\%, and only the maximum depenetration velocity, the parameter PhysX's documentation names for bodies that start in overlap~\citep{physx54}, matters: limited to 1\,m/s, the removal stretches to 16.5\,ms and the median distance falls to 10\,cm; at 0.1\,m/s no pose throws the can. MuJoCo removes the same overlaps over 38.6\,ms and throws the can in 5\% of poses. Bullet~\citep{coumans2016pybullet}, a third engine, pushes the can out of a 20\,mm overlap at 1.02\,m/s by default, close to MuJoCo's 1.05\,m/s, and at 10.22\,m/s with its error correction raised to 0.8 (\cref{app:stress}).

Replays of recorded episodes need the second form of rerun, since a change from the first step alters outcomes that no artifact decided (\cref{app:reruns}). Switched on 0.1\,s before the first artifact, with everything before identical to the recording, only the depenetration limit keeps the object slow (\cref{tab:cause-replay}): the can's median peak speed after the switch is 1.39\,m/s, against 3.72\,m/s with a ten times heavier can and 11.20\,m/s with more solver iterations, and the two rescue fewer failures, 14 and 12 of 21 against 18.

\subsection{What the artifacts do to the score}
\label{sec:exp-fix}

Every episode with an artifact is replayed with its recorded actions and the depenetration limit of 1\,m/s on the ejected body, from 0.1\,s before to 0.2\,s after the first artifact; the benchmark's success check scores the replay, and \cref{eq:corrected} replaces the recorded outcome. All 69 replays on the five tasks match their recordings bit for bit up to the switch. \Cref{app:manifest} lists every replayed episode. Of LingBot-VA's 40 failures with an artifact on the five tasks, 21 pass; of its 19 successes with an artifact, 10 fail (\cref{tab:scores}). Its success rate on the five tasks moves from 83.1\% to 87.3\%. HyVLA gains 4 failures, loses 3 successes, and moves from 90.0\% to 91.0\%. Over all 50 tasks, 2060 episodes, LingBot-VA moves from 90.8\% to 93.0\% under exclusion. As a control, the same limit applied to 133 episodes without an artifact changes no outcome. In the repaired engine, new rollouts on the five tasks succeed in 98 of 100 for LingBot-VA against 89, and in 94 for HyVLA against 90, with 0 and 1 artifacts against 12 and 10 (\cref{tab:repaired}). On 39 other tasks, 20 episodes each, the repaired engine leaves LingBot-VA's rate unchanged: 719 of 780 against 721. Across five models, the released engine ranks HyVLA first, while the repaired engine gives the same order as a real robot (\cref{app:rank}).

A benchmark could instead drop every confirmed flagged episode. That gives 98.0\% for LingBot-VA and 95.6\% for HyVLA on the five tasks (\emph{Excl.}, \cref{tab:scores}). Keeping the limit on for whole rollouts changes the task (\emph{CL}): LingBot-VA reaches 87.1\% on the five tasks against 81.8\% as run on the same seeds, but Place Object Basket, with two artifacts in 20 episodes, falls from 90.0\% to 50.0\%. HyVLA falls from 90.7\% on the shared seeds to 78.4\%, to 70.0\% on Place Bread Basket and 45.5\% on Place Object Basket. We use the replay rate as the point estimate (\cref{app:narrow}).

\begin{table}[t]
  \centering
  \caption{\textbf{Closed loop evaluation in the repaired engine.} Successes of 20; episodes with an artifact in parentheses. \emph{Bench.}: RoboTwin as released. \emph{Offset}: contact offset 1\,mm. \emph{Repaired}: the smaller threshold.}
  \label{tab:repaired}
  \small
  \begin{tabular}{@{}lccccc@{}}
    \toprule
    & \multicolumn{3}{c}{LingBot-VA} & \multicolumn{2}{c}{HyVLA} \\
    \cmidrule(lr){2-4}\cmidrule(l){5-6}
    Task & Bench. & Offset & Repaired & Bench. & Repaired \\
    \midrule
    Place Can Basket    & 15 (5) & 19 (3) & 20 (0) & 16 (4) & 18 (0) \\
    Stack Bowls Three   & 15 (6) & 17 (3) & 19 (0) & 18 (3) & 20 (0) \\
    Place Object Basket & 20 (0) & 20 (0) & 20 (0) & 19 (0) & 18 (0) \\
    Place Bread Basket  & 19 (1) & 19 (0) & 20 (0) & 20 (0) & 18 (1) \\
    Stack Bowls Two     & 20 (0) & 20 (0) & 19 (0) & 17 (3) & 20 (0) \\
    \midrule
    Five tasks, of 100  & 89 (12) & 95 (6) & 98 (0) & 90 (10) & 94 (1) \\
    \bottomrule
  \end{tabular}
\end{table}

\subsection{Other platforms}
\label{sec:exp-platforms}

\begin{table}[t]
  \centering
  \caption{\textbf{Four benchmarks, two engines.} Episodes with a contact artifact, the fastest object speed at an artifact, and the speed when replayed with the depenetration limit of 1\,m/s.}
  \label{tab:platforms}
  \small
  \setlength{\tabcolsep}{4pt}
  \begin{tabular*}{\linewidth}{@{\extracolsep{\fill}}lllcccc@{}}
    \toprule
    Benchmark & Engine & Policy & Episodes & With artifact & Peak (m/s) & Limited (m/s) \\
    \midrule
    RoboTwin~2.0 & PhysX, SAPIEN    & LingBot-VA  & 2060 & 80 & 21.8 & 1.39 \\
                 &                  & HyVLA       & 220  & 10 & --   & -- \\
    RoboDojo     & PhysX, Isaac Sim & XR-1        & 40   & 7  & 6.30 & 1.88 \\
    LIBERO       & MuJoCo           & $\pi_{0.5}$ & 400  & 2  & 1.57 & -- \\
    RoboCasa     & MuJoCo           & XR-1        & 300  & 5  & 1.19 & -- \\
    \bottomrule
  \end{tabular*}
\end{table}

\begin{table}[t]
  \centering
  \caption{\textbf{Every benchmark we audited.} Success rate in \%, as reported and after removing episodes with a real artifact.}
  \label{tab:all-bench}
  \small
  \begin{tabular}{@{}lllccc@{}}
    \toprule
    Benchmark & Engine & Policy & Episodes & Reported & Corrected \\
    \midrule
    RoboTwin~2.0 & PhysX, SAPIEN & LingBot-VA & 2060 & 90.8 & 93.0 \\
                 &               & HyVLA      & 220  & 78.2 & 80.0 \\
    RoboLab      & PhysX, Isaac Sim & $\pi_{0.5}$-DROID & 2400 & 28.1 & 28.5 \\
                 &                  & GR00T~N1.7-DROID  & 2400 & 8.8  & 9.1 \\
    RoboDojo     & PhysX, Isaac Sim & XR-1              & 40   & 40.0 & 48.5 \\
                 &                  & InternW0          & 360  & 39.7 & 40.6 \\
                 &                  & OpenWAM           & 360  & 19.4 & 19.8 \\
                 &                  & $\pi_{0.5}$       & 360  & 17.1 & 17.0 \\
                 &                  & X-VLA             & 360  & 15.8 & 16.7 \\
                 &                  & Xiaomi-Robotics-0 & 360  & 5.2  & 5.3 \\
                 &                  & $\pi_0$           & 360  & 3.6  & 3.7 \\
                 &                  & InternVLA-A1      & 360  & 2.1  & 2.1 \\
                 &                  & GR00T~N1.7        & 360  & 1.3  & 1.3 \\
    LIBERO       & MuJoCo & $\pi_{0.5}$ & 400 & 95.2 & 95.5 \\
    RoboCasa     & MuJoCo & XR-1        & 300 & 61.7 & 62.7 \\
    \bottomrule
  \end{tabular}
\end{table}

\sg{} also attaches to RoboLab~\citep{yang2026robolab}, RoboDojo, LIBERO and RoboCasa (\cref{tab:platforms}, \cref{tab:all-bench}). On RoboLab, it flags 20.7\% and 11.2\% of the 2400 episodes of $\pi_{0.5}$-DROID and GR00T~N1.7-DROID; after human review, their success rates move from 28.1\% to 28.5\% and from 8.8\% to 9.1\%. On RoboDojo, the correction changes no pair's order among eight models with real robot results; their Spearman correlation with the real success rate stays 0.81 (\cref{app:robodojo-real}). RoboDojo, on PhysX, ejects objects as RoboTwin does: 7 of 40 XR-1 episodes contain artifacts, all failures, and the same limit slows a tube from 4.31 to 1.88\,m/s. On MuJoCo, artifacts are few and slow: 2 of 400 LIBERO episodes and 5 of 300 RoboCasa episodes, none above 1.6\,m/s.

\subsection{Two other simulation errors}
\label{sec:exp-other}

\paragraph{Objects that start inside their support.}
On LIBERO, 650 of the 2000 stored initial states place objects inside the floor or the table, in all 10 LIBERO-Object tasks and in 3 LIBERO-10 tasks (\cref{tab:libero-init}). Objects sit up to 37.7\,mm deep, and MuJoCo pushes them out at up to 1.83\,m/s with no robot contact. They come to rest before the policy acts, and the score barely changes: $\pi_{0.5}$ succeeds in 96.2\% of the 130 episodes that start from these states, against 94.8\% in the other 270.

\begin{figure}[t]
  \centering
  \includegraphics[width=\linewidth]{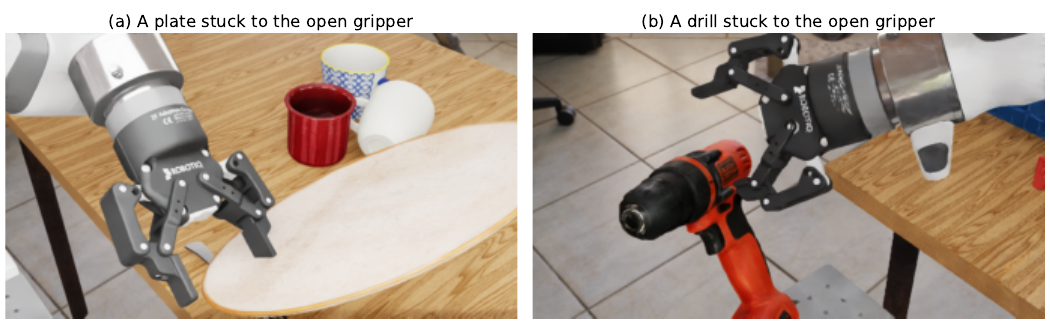}
  \caption{\textbf{Objects stuck to an open gripper} in two GR00T~N1.7 episodes on RoboLab. The gripper is fully open and the object touches only one finger. (a) The plate has risen 10\,cm with the finger. (b) The drill hangs from the finger past the table edge.}
  \label{fig:robolab-stick}
\end{figure}

\paragraph{Objects stuck to an open gripper.}
On RoboLab, a rule flags an episode if, 0.5\,s after the gripper opens, the object touches only the fingers and moves 10\,cm or more with them within 2\,s. It flags 179 of the 4800 episodes (3.7\%). In 10 of 20 flagged clips we checked by eye, the object sticks to the gripper or passes through it (\cref{tab:robolab-stick,fig:robolab-stick}). Dropping the flagged episodes moves the score of $\pi_{0.5}$ from 28.1\% to 29.2\% and that of GR00T~N1.7 from 8.8\% to 9.1\%.

\subsection{Reliability of replay and detection}
\label{sec:exp-reliability}

\sg{} reproduces every recorded episode exactly, and only with the replay details of \cref{sec:method-replay} (\cref{tab:reliability}).

Detection agrees with human judgment: five robotics engineers rated each clip independently, without knowing where it came from, and settled disagreements by discussion (\cref{tab:reliability}, \cref{app:labels}). On Place Can Basket, they judged 37 of 40 clips centered on detected events physically implausible (precision 0.925) and none of 20 hard negatives. On RoboLab and RoboDojo, 26\% and 30\% of 200 checked flags are real artifacts, and 1 of 200 checked unflagged episodes holds a missed one (\cref{app:review}). The kind of flag tells how far to trust it (\cref{tab:review}). On RoboDojo, contacts shallower than 3\,mm are real in 8 of 62 checked flags, and the two deep kinds in 24 of 38; keeping only the deep kinds raises the precision from 30\% to 65\% and keeps 71\% of the real artifacts. On RoboLab, the shelf rack and the fruit that starts inside its plate give 1 real artifact in 39 checked flags.

\section{Discussion and Limitations}
\label{sec:discussion}

\paragraph{Discussion.}
The four scores of \cref{tab:scores} answer four questions about the same episodes. The benchmark's own scripted expert, acting on ground truth poses, fails in the same way and is rescued by the same limit (\cref{tab:expert}). The reported overlap appears within one step at a resting contact with the thin wall; its one step removal is the artifact because of what it produces: a can drifting at 0.2\,m/s gains 2.4\,J while no robot link touches it (\cref{fig:ejection}). A limit kept on for whole rollouts changes the task; the smaller threshold removes the false contact (\cref{tab:repaired}). As reported, HyVLA leads LingBot-VA on the five tasks, 90.0 against 83.1; under exclusion the order reverses, 95.6 against 98.0. New rollouts in the repaired engine give 94 against 98 of 100, a gap within sampling noise: on 200 more rollouts in the repaired engine, HyVLA succeeds in 196 (\cref{app:cause}).

\paragraph{Limitations.}
The main experiments are on RoboTwin; on the four other benchmarks we did not rerun the cause. On RoboDojo, PhysX on the GPU also reports false contacts that we have not yet traced. The smaller threshold removes one cause; objects that stick to the gripper or pass through it on RoboLab have another, which we have not repaired. The detector selects episodes for a person to check: on RoboLab and RoboDojo, fewer than a third of its flags are real, so the scores we report there use the human review.

\section{Conclusion}
\label{sec:conclusion}

Success rate in simulation is read as a policy's ability to act, but it also counts the failures the simulator causes. On tasks that place objects into containers with thin walls, most failures of two policies and of the benchmark's own scripted demonstrations contain a contact artifact. Controlled reruns trace the artifact to a false contact from a fixed threshold in PhysX's collision detection, and a smaller threshold removes it: in the repaired engine, LingBot-VA succeeds in 98 of 100 new rollouts on these tasks against 89, and its rate on 39 other tasks does not change. HyVLA rises from 269 to 290 of 300 rollouts, and its episodes with an artifact fall from 40 to 1. Audits of four further benchmarks find artifacts on both PhysX and MuJoCo. On RoboLab, the audit finds two more classes: objects that stick to an open gripper, and objects that pass through it. Replaying the recorded actions with the removal slowed moves LingBot-VA's rate on the five tasks from 83.1\% to 87.3\%. An audit of this kind lets a benchmark separate what the policy did from what the solver did.

We suggest three steps for a benchmark. First, record the state at every physics step, so that any episode can be replayed and checked. Second, report with each success rate the number of episodes that contain an artifact, and the rate without them. Third, when an artifact is frequent, trace it to the engine and repair it there; the repaired engine then gives the score directly, and no correction is needed. \sg{} supports these steps on SAPIEN, Isaac Sim and MuJoCo. Its detector reads only the body states and contacts at each step, so a new engine needs only an adapter. The audit runs inside the benchmark's own evaluation script and leaves its protocol unchanged.

\subsection*{AI use statement}

In this work, we have not used generative AI tools to generate synthetic data, to develop theoretical models, to formulate or prove mathematical claims, or to propose hypotheses, and translation, dataset cleaning and qualitative analysis are not applicable to this work. We used generative AI tools to create and edit software code, to draft and edit parts of the paper, to identify relevant literature and to format references. We have reviewed all AI-assisted work: every number in the paper comes from runs the authors executed and can be regenerated by exact replay. We take responsibility for the final content of this work, including text, claims or artifacts produced with the aid of generative AI.

\subsection*{Ethics statement}

This work studies the validity of simulated robot benchmarks. It involves no personal data; the human judgments in \cref{app:labels} are annotations of simulation videos by five robotics engineers.

\subsection*{Reproducibility statement}

\sg{} runs the benchmark's own evaluation script, and every episode in this paper is stored with its seed and the actuation of every physics step, so each result can be regenerated by exact replay. \Cref{app:runs} lists tasks, episodes, upstream versions and checkpoints; \cref{app:detector} lists every detector parameter; \cref{app:impl} describes the adapters and the replay; \cref{app:stress} specifies the stress test. The code, the detector, the replay and rerun tools, the episode manifest of \cref{app:manifest} and the scripts that produce each table are at \url{https://github.com/ALEXander22568/simuguard}; the recordings will be released with the code.

\bibliography{references}
\bibliographystyle{iclr2027_conference}

\appendix
\raggedbottom
\section{Extended Related Work}
\label{app:related}

\paragraph{Policies.}
Language conditioned manipulation moved from affordances~\citep{ahn2022saycan} and multimodal prompts~\citep{jiang2023vima} to transformers trained on large robot data~\citep{brohan2023rt1,oxe2024,octo2024}, then to VLAs built on a vision language backbone~\citep{driess2023palme,brohan2023rt2,li2024roboflamingo,kim2024openvla,black2024pi0,intelligence2025pi05,bjorck2025gr00t,geminirobotics2025}. Later work changes the action head~\citep{chi2023diffusionpolicy,zhao2023act,ke2024diffuseractor,liu2025rdt,pertsch2025pi0fast,kim2025openvlaoft} or the spatial input~\citep{zheng2025tracevla,qu2025spatialvla}, trained on large real robot datasets~\citep{walke2023bridgev2,khazatsky2024droid}. World action models also predict future observations: video prediction as a planner~\citep{du2023unipi,ko2024avdc}, video generative pretraining~\citep{wu2024gr1,cheang2024gr2,hu2025vpp}, one model for video and actions~\citep{li2025uva,li2026lingbotva}, and latent actions learned from video~\citep{ye2025lapa}. They build on world models for control~\citep{wu2022daydreamer,hafner2025dreamerv3} and interactive video models~\citep{yang2024unisim,bruce2024genie}. A further line uses a coding agent as the policy: the agent writes, calls and evolves robot programs~\citep{gao2026selfevolving,ge2026codeaspolicy,gao2026hexashell}, and EmbodiedLab benchmarks such agents on laboratory experiments~\citep{gao2026embodiedlab}. Simulation scores all of them by episode outcome, which \sg{} checks.

\paragraph{Simulated benchmarks.}
Manipulation suites include RLBench~\citep{james2020rlbench}, Meta-World~\citep{yu2019metaworld}, robosuite~\citep{zhu2020robosuite}, CALVIN~\citep{mees2022calvin}, ManiSkill~\citep{mu2021maniskill1,gu2023maniskill2,tao2024maniskill3}, LIBERO~\citep{liu2023libero}, RoboCasa~\citep{nasiriany2024robocasa}, RoboTwin~\citep{mu2025robotwin,chen2025robotwin2}, RoboDojo~\citep{chen2026robodojo} and NVIDIA RoboLab~\citep{yang2026robolab}. Household suites include iGibson~\citep{li2021igibson2}, Habitat~\citep{szot2021habitat2} and BEHAVIOR~\citep{li2022behavior1k}. GPU physics~\citep{makoviychuk2021isaacgym,mittal2023orbit} makes large evaluations cheap. Generalization suites vary objects, layouts, lighting and instructions~\citep{pumacay2024colosseum,garcia2025gembench,zhou2025liberopro,fei2025liberoplus}. SimplerEnv~\citep{li2024simplerenv} and RoboArena~\citep{atreya2025roboarena} check rankings against real robots, and \citet{acosta2022validating} validate engines on recorded impacts. All of them treat the physics engine as fixed.

\paragraph{Evaluation practice.}
Work on robot evaluation covers trial counts, conditions and uncertainty~\citep{kressgazit2024evaluation,vincent2024generalizable,snyder2026binary}. RoboEval~\citep{wang2025roboeval} and \citet{jiang2026benchmarking} ask what manipulation benchmark scores measure. In machine learning, seeds and variance distort comparisons~\citep{henderson2018matters,agarwal2021precipice,bouthillier2021variance}. These works improve how outcomes are pooled. \sg{} asks whether each outcome is the policy's.

\paragraph{Contact simulation.}
Engines model contact differently: soft constraints with a time constant in MuJoCo~\citep{todorov2012mujoco}, hard constraints with penetration correction in PhysX, compliant and pressure field models~\citep{castro2023convex,elandt2019hydroelastic}, and differentiable formulations~\citep{howell2022dojo}. Comparisons across engines~\citep{erez2015simulation,lelidec2023contact} and against real impacts~\citep{acosta2022validating} show that one scene yields different forces and motion. Convex decomposition of assets adds another approximation~\citep{wei2022coacd}. \sg{} finds contacts that an engine reports but the geometry lacks, and measures their effect on scores.

\paragraph{Faults in physics engines.}
PhysX's documentation warns that bodies starting in overlap are pushed apart fast, and suggests limiting the depenetration velocity~\citep{physx54}. It does not say that the engine can report an overlap that does not exist, or how this affects a benchmark. PhyFu~\citep{xiao2023phyfu} fuzzes physics engines for results that break physical laws. PhysiXFails~\citep{li2025physixfails} finds them in software built on physics engines, such as games. Neither looks at robot policy scores. A developer survey names unrealistic physics as a main obstacle to testing robots in simulation~\citep{afzal2021simulation}. Reinforcement learning agents exploit simulator bugs during training~\citep{baker2020emergent,krakovna2020specification}. We study evaluation, where a bug changes the outcome of an episode.

\section{Setup}
\label{app:runs}

\subsection{RoboTwin}
We use RoboTwin commit 6dde571, XPolicyLab commit fa431ec, LingBot-VA code 7c6ffa9, and the RoboTwin checkpoint of LingBot-VA from ModelScope with end effector actions. Inference runs on RTX~4090 or H800 GPUs with identical weights and launch settings. RoboTwin skips each seed whose scripted demonstration fails its check, until $N$ seeds are scored. Every run uses the \emph{demo\_clean} configuration, seen instructions, seed base 0 and the expert check. The 40 tasks outside the cross task study ran in two passes of 20 seeds from the same seed base. The second pass ran on H800 GPUs with RoboTwin commit 2a14726 and XPolicyLab fe71eb5. The passes share 675 seeds and agree on the outcome of 629 of them. We replay only the first pass. Of the ten cross task tasks, Place Can Basket has 90 episodes and Stack Bowls Three has 50. The other eight have 40 each, in two replayed passes on RTX~4090 GPUs. The three replays of Put Object Cabinet's second pass diverge from the recording before the switch, so \cref{tab:manifest} lists them but we do not count them. Place Can Basket ran three times with 30 scored seeds each, because one run is not stable: 15 of its 37 seeds succeed in some runs and fail in others. The runs give 23, 24 and 20 successes, or $74.4 \pm 6.9\%$. \Cref{tab:runs} lists every task.

\begin{tableH}
  \centering
  \caption{\textbf{RoboTwin~2.0 runs} of LingBot-VA. \emph{Artifact}: episodes with a contact artifact, among failures and successes. \emph{Skipped}: seeds dropped because the scripted demonstration failed. \emph{Succ.}: successes. \emph{Rep.}: reported rate. Bold: the five tasks of \cref{tab:scores}. The 40 tasks outside the cross task study pool two passes of 20 seeds.}
  \label{tab:runs}
  \scriptsize\setlength{\tabcolsep}{2pt}
  \begin{tabular}{@{}lcccccc@{}}
    \toprule
    & & & & \multicolumn{2}{c}{Artifact in} & \\
    \cmidrule(lr){5-6}
    Task & $N$ & Succ. & Rep. (\%) & fail. & succ. & Skipped \\
    \midrule
    Adjust Bottle & 40 & 40 & 100.0 & 0 & 0 & 11 \\
    Beat Block Hammer & 40 & 39 & 97.5 & 0 & 0 & 19 \\
    Blocks Ranking RGB & 40 & 39 & 97.5 & 0 & 0 & 0 \\
    Blocks Ranking Size & 40 & 36 & 90.0 & 0 & 0 & 3 \\
    Click Alarmclock & 40 & 40 & 100.0 & 0 & 0 & 8 \\
    Click Bell & 40 & 40 & 100.0 & 0 & 0 & 0 \\
    Dump Bin Bigbin & 40 & 39 & 97.5 & 1 & 0 & 36 \\
    Grab Roller & 40 & 40 & 100.0 & 0 & 0 & 2 \\
    Handover Block & 40 & 39 & 97.5 & 0 & 0 & 12 \\
    Handover Mic & 40 & 38 & 95.0 & 0 & 0 & 6 \\
    Hanging Mug & 40 & 9 & 22.5 & 1 & 0 & 7 \\
    Lift Pot & 40 & 40 & 100.0 & 0 & 0 & 56 \\
    Move Can Pot & 40 & 37 & 92.5 & 0 & 0 & 4 \\
    Move Pillbottle Pad & 40 & 40 & 100.0 & 0 & 0 & 21 \\
    Move Playingcard Away & 40 & 40 & 100.0 & 0 & 0 & 0 \\
    Move Stapler Pad & 40 & 27 & 67.5 & 2 & 0 & 2 \\
    Open Laptop & 40 & 40 & 100.0 & 0 & 0 & 0 \\
    Open Microwave & 40 & 31 & 77.5 & 0 & 0 & 32 \\
    Pick Diverse Bottles & 40 & 38 & 95.0 & 0 & 0 & 46 \\
    Pick Dual Bottles & 40 & 40 & 100.0 & 0 & 0 & 2 \\
    Place A2B Left & 40 & 40 & 100.0 & 0 & 0 & 9 \\
    Place A2B Right & 40 & 37 & 92.5 & 0 & 0 & 9 \\
    \textbf{Place Bread Basket} & 40 & 39 & 97.5 & 1 & 0 & 9 \\
    Place Bread Skillet & 40 & 39 & 97.5 & 0 & 0 & 56 \\
    Place Burger Fries & 40 & 40 & 100.0 & 0 & 5 & 0 \\
    \bottomrule
  \end{tabular}\hfill
  \begin{tabular}{@{}lcccccc@{}}
    \toprule
    & & & & \multicolumn{2}{c}{Artifact in} & \\
    \cmidrule(lr){5-6}
    Task & $N$ & Succ. & Rep. (\%) & fail. & succ. & Skipped \\
    \midrule
    \textbf{Place Can Basket} & $3\times30$ & 67 & 74.4 & 21 & 16 & 22 \\
    Place Cans Plasticbox & 40 & 40 & 100.0 & 0 & 0 & 0 \\
    Place Container Plate & 40 & 39 & 97.5 & 0 & 0 & 9 \\
    Place Dual Shoes & 40 & 37 & 92.5 & 3 & 2 & 27 \\
    Place Empty Cup & 40 & 40 & 100.0 & 0 & 0 & 4 \\
    Place Fan & 40 & 36 & 90.0 & 1 & 0 & 8 \\
    Place Mouse Pad & 40 & 39 & 97.5 & 0 & 0 & 0 \\
    \textbf{Place Object Basket} & 40 & 35 & 87.5 & 3 & 3 & 16 \\
    Place Object Scale & 40 & 36 & 90.0 & 0 & 0 & 16 \\
    Place Object Stand & 40 & 40 & 100.0 & 0 & 0 & 1 \\
    Place Phone Stand & 40 & 39 & 97.5 & 0 & 0 & 20 \\
    Place Shoe & 40 & 39 & 97.5 & 0 & 0 & 3 \\
    Press Stapler & 40 & 33 & 82.5 & 0 & 0 & 4 \\
    Put Bottles Dustbin & 40 & 29 & 72.5 & 0 & 0 & 15 \\
    Put Object Cabinet & 40 & 30 & 75.0 & 1 & 2 & 272 \\
    Rotate QRcode & 40 & 37 & 92.5 & 0 & 0 & 3 \\
    Scan Object & 40 & 38 & 95.0 & 0 & 0 & 80 \\
    Shake Bottle & 40 & 40 & 100.0 & 0 & 0 & 16 \\
    Shake Bottle Horizontally & 40 & 40 & 100.0 & 0 & 0 & 16 \\
    Stack Blocks Three & 40 & 35 & 87.5 & 2 & 0 & 3 \\
    Stack Blocks Two & 40 & 40 & 100.0 & 0 & 0 & 0 \\
    \textbf{Stack Bowls Three} & 50 & 38 & 76.0 & 12 & 0 & 43 \\
    \textbf{Stack Bowls Two} & 40 & 37 & 92.5 & 3 & 0 & 15 \\
    Stamp Seal & 40 & 38 & 95.0 & 1 & 0 & 18 \\
    Turn Switch & 40 & 21 & 52.5 & 0 & 0 & 7 \\
    \bottomrule
  \end{tabular}
  \par\vspace{2pt}\noindent All 50 tasks: $N$ = 2060, successes 1870 (90.8\%), artifacts in 52 failures and 28 successes, 968 seeds skipped.
\end{tableH}

\paragraph{HyVLA.}
We run Hy-Embodied-0.5-VLA-RoboTwin through XPolicyLab with its own RoboTwin settings: end effector actions, seven actions per query. The evaluator arguments match those of LingBot-VA, so both policies are scored on the same seed sequence. Place Object Basket ran as five parallel shards with different start seeds. The demonstration check can pass on a seed in one run and fail in another, so two shards scored two seeds twice. We keep one run per seed, leaving 20 seeds (\cref{tab:runs-hyvla}).

\begin{tableH}
  \centering
  \caption{\textbf{RoboTwin~2.0 runs} of HyVLA. \emph{Artifact}: failures that contain a contact artifact. \emph{Skipped}: seeds dropped because the scripted demonstration failed.}
  \label{tab:runs-hyvla}
  \small
  \begin{tabular}{lccccc}
    \toprule
    Task & $N$ & Successes & Reported (\%) & Artifact & Skipped \\
    \midrule
    \textbf{Place Can Basket}    & 20 & 16 & 80.0  & 1/4 & 5 \\
    \textbf{Stack Bowls Three}   & 20 & 18 & 90.0  & 2/2 & 18 \\
    \textbf{Place Object Basket} & 20 & 19 & 95.0  & 0/1 & 8 \\
    \textbf{Place Bread Basket}  & 20 & 20 & 100.0 & 0/0 & 5 \\
    \textbf{Stack Bowls Two}     & 20 & 17 & 85.0  & 3/3 & 7 \\
    Blocks Ranking Size          & 20 & 17 & 85.0  & 0/3 & 2 \\
    Open Microwave               & 20 & 9  & 45.0  & 0/11 & 2 \\
    Put Bottles Dustbin          & 20 & 15 & 75.0  & 0/5 & 10 \\
    Put Object Cabinet           & 20 & 16 & 80.0  & 0/4 & 156 \\
    Stamp Seal                   & 20 & 15 & 75.0  & 0/5 & 9 \\
    Turn Switch                  & 20 & 10 & 50.0  & 0/10 & 3 \\
    \bottomrule
  \end{tabular}
\end{tableH}

\subsection{LIBERO}
We use upstream LIBERO with robosuite~1.4.0 and MuJoCo~3.2.3, a 2\,ms physics step and 25 physics steps per action. The policy is the $\pi_{0.5}$ checkpoint RLinf-Pi05-LIBERO-130-fullshot-SFT~\citep{yu2025rlinf} with its training protocol: two camera images and the robot state in, five actions per query out. The arm rests for 15 steps before the policy acts, an episode has at most 600 policy steps, and it stops at the first success. We run initial states 0 to 9 of all 40 tasks, with the seed equal to the state index. The policy succeeds in 381 of 400 episodes (95.2\%): 99 on LIBERO-Spatial, 97 on LIBERO-Object, 98 on LIBERO-Goal and 87 on LIBERO-10.

\subsection{RoboCasa}
We use RoboCasa~\citep{nasiriany2024robocasa} on MuJoCo~3.3.1 with robosuite's split physics step, a 2\,ms physics step and 25 physics steps per action. We run XR-1~\citep{fan2025xr1} with its RoboCasa checkpoint and evaluation protocol. Ten object placement tasks get ten episodes each: Arrange Bread Basket, Coffee Setup Mug, Pick Place Counter To Cabinet, Pick Place Counter To Stove, Pick Place Drawer To Counter, Pick Place Sink To Counter, Pick Place Toaster To Counter, Stack Bowls Cabinet, Store Leftovers In Bowl and Wash Fruit Colander. The other 40 tasks of XR-1's evaluation set get five each. The set has 18 atomic and 32 composite tasks, and 16 of them are unseen in XR-1's training. The ten placement tasks are 4 atomic and 6 composite. XR-1 succeeds in 72 of the first 100 episodes and in 113 of the other 200.

\subsection{RoboDojo}
RoboDojo~\citep{chen2026robodojo} runs Isaac Sim~5.1.0 and Isaac Lab~0.54.3 with two ARX~X5 arms, a 4\,ms physics step, 10 physics steps per action and PhysX GPU dynamics. We run XR-1 with its official RoboDojo checkpoint on four tasks: Fill Pen Holder (three layouts), Insert Tubes (three), Put Bottles Into Dustbin (two) and Store Tools In Toolbox (two), with the benchmark's three evaluation seeds. Seed 0 ran twice: once with every episode replayed, and once without replay together with seeds 1 and 2. XR-1 succeeds in 16 of the 40 episodes.

\subsection{Detector parameters}
\label{app:detector}
\begin{tableH}
  \centering
  \caption{\textbf{Detector parameters} (\cref{sec:method-detect}), the same for all tasks and engines. Velocity change thresholds are for a 4\,ms step and scale with the physics step on other engines.}
  \label{tab:detector}
  \small
  \begin{tabular}{llc}
    \toprule
    Stage & Parameter & Value \\
    \midrule
    Contact ejection & recent contact window & 40\,ms \\
     & minimum speed & 0.5\,m/s \\
     & minimum velocity change in one 4\,ms step & 0.35\,m/s \\
     & confirmation window & 240\,ms \\
     & minimum displacement & 4\,cm \\
     & free flight & $\ge$ 16\,ms \\
     & or violent contact with the container: velocity change / speed & $\ge$ 0.75 / 1.0\,m/s \\
    Gravity & window before onset for the drop height $\Delta z$ & 0.6\,s \\
     & window after onset for the peak speed & 0.1\,s \\
     & ratio of peak speed to $\sqrt{2g\,\Delta z}$ & $> 1.5$ \\
     & speed floor in the denominator & 0.05\,m/s \\
    Carrier & window before onset for carriers & 40\,ms \\
     & window after onset for the peak speed & 0.1\,s \\
     & ratio of peak speed to the fastest carrier's speed & $\ge 1.5$ \\
     & speed floor in the denominator & 0.05\,m/s \\
    \bottomrule
  \end{tabular}
\end{tableH}

\section{Reliability of the Tool}
\label{app:reliability}

\subsection{Recording, replay and reruns}
\label{app:impl}
\paragraph{Recording.}
\sg{} wraps the physics step of the running scene and leaves the evaluator, benchmark and policy unchanged. In SAPIEN, it records the drive targets, drive velocity targets and joint forces of every articulation before each step. In MuJoCo, it records control inputs, applied forces, mocap poses, the solver's warm start when benchmark code changed it, and every write to the state between steps. In Isaac Sim, it records the same articulation inputs as in SAPIEN. It reads states from PhysX tensor views and contacts from the PhysX contact report, enabled in RoboDojo's object loaders. Body velocities are those of the centers of mass. \Cref{tab:reliability} gives the recording cost on each engine.

\paragraph{Replay.}
A replay rebuilds the scene from its seed in a fresh process and applies the recorded actuation from the first step. In RoboTwin, it also restores the commanded gripper opening at each step, because the success checks read it. In MuJoCo, it writes back the recorded warm start and the writes between steps, and checks the full integration state against snapshots saved every 250 steps. LIBERO on robosuite~1.4.0 places fixtures with an unseeded random generator, so the seed does not fix the scene. \sg{} therefore writes the recorded body poses back before the replay. In Isaac Sim, restoring a saved state in place leaves trajectories 0.2 to 1\,mm apart after 500 steps, because PhysX's contact caches cannot be restored. Rebuilding the scene and replaying the recorded actuation is exact.

\paragraph{Reruns.}
The depenetration limit applies to every dynamic body and articulation link until the episode ends. A rerun switches it on 0.1\,s (25 physics steps) before the onset of the episode's first contact artifact. It checks every state before that step against the recording.

\paragraph{Replay checks, detection checks and recording cost.}
\Cref{tab:reliability} lists every replay check, the detection checks against human judgment, and the recording cost on each engine.

\begin{tableH}
  \centering
  \caption{\textbf{Reliability of \sg{} and cost of recording.} \emph{Exact}: every recorded float of every tracked body matches the recording.}
  \label{tab:reliability}
  \footnotesize
  \begin{tabular}{@{}>{\raggedright\arraybackslash}p{0.56\linewidth}>{\raggedright\arraybackslash}p{0.15\linewidth}>{\raggedright\arraybackslash}p{0.23\linewidth}@{}}
    \toprule
    Check & Platform & Result \\
    \midrule
    \emph{Replay} & & \\
    Fresh process, 96 failures (2.42\,M steps); events and outcomes reproduced & RoboTwin & 96/96 exact; 43/43, 96/96 \\
    Stack Bowls Three, 12 failures: outcomes with / without gripper restore & RoboTwin & 12/12 / 7/12 \\
    Another machine, 25 episodes & RoboTwin & 25/25 exact \\
    Prefix before a rerun switch, 273 replays & RoboTwin & 273/273 exact \\
    Full episodes, 400 (1.88\,M steps); without the recorded warm start & LIBERO & 400/400 exact; 0/33 \\
    Rebuilt scene, 13 episodes & RoboDojo & 13/13 exact \\
    Monitor on versus off & RoboTwin, RoboDojo & 59 and 26 bodies identical \\
    State restored in place during contact & SAPIEN / Isaac Sim & mm to cm / 0.2--1\,mm apart \\
    \midrule
    \emph{Detection} & & \\
    Detected events judged implausible (Place Can Basket) & RoboTwin & 37/40 \\
    Hard negatives judged implausible (Place Can Basket) & RoboTwin & 0/20 \\
    Agreement on 60 episodes of six tasks, with / without the gravity stage & RoboTwin & 58/60 / 39/60 \\
    \midrule
    \emph{Recording cost per physics step} & SAPIEN & 7.5--11.6\,ms, 85--105 bytes \\
     & MuJoCo (LIBERO) & 2.8\,ms, 659 bytes, +23\% wall time \\
     & Isaac Sim & 6.9--18.4\,ms, 3--7\% of episode wall time \\
    \bottomrule
  \end{tabular}
\end{tableH}

\subsection{What each detection stage removes}
\Cref{tab:stages} counts the events each stage leaves on each platform. The gravity stage matters most where objects are dropped. On RoboTwin's Put Bottles Dustbin, it removes all 20 flagged drops. On 60 human-labeled episodes of six more RoboTwin tasks, it raises agreement with the annotators from 39 to 58 (\cref{tab:reliability}). The carrier stage changes no label on RoboTwin and RoboDojo, and removes none of RoboCasa's remaining events.

\begin{tableH}
  \centering
  \caption{\textbf{Detection stages across platforms.} Events flagged by the contact ejection stage and left after the gravity and carrier stages. Episodes in parentheses where counted.}
  \label{tab:stages}
  \small
  \begin{tabular}{@{}llccc@{}}
    \toprule
    Platform & Episodes & Flagged & After gravity & After carrier \\
    \midrule
    RoboTwin, Put Bottles Dustbin (LingBot-VA) & 10 & 20 & 0 & 0 \\
    RoboTwin, 40 tasks, second pass (LingBot-VA) & 800 & 12 (12) & 10 (10) & 10 (10) \\
    RoboTwin, eight tasks, second pass (LingBot-VA) & 200 & 169 (58) & 22 (10) & 17 (10) \\
    RoboCasa, ten placement tasks (XR-1) & 100 & 7 (7) & 3 (3) & 3 (3) \\
    RoboCasa, the other 40 tasks (XR-1) & 200 & 15 (15) & 2 (2) & 2 (2) \\
    RoboDojo, four tasks (XR-1) & 40 & 25 (15) & 12 (7) & 12 (7) \\
    LIBERO, stored initial states, no policy & 2000 & 655 & 650 & 650 \\
    \bottomrule
  \end{tabular}
\end{tableH}

\section{RoboTwin: Cause, Stress Test and Reruns}

\subsection{Cause in the collision detection}
\label{app:cause}

\paragraph{How \sg{} finds the cause.}
We found the cause in three steps. First, the recordings show the can and the basket wall apart one step before each artifact and deeply overlapped one step later. They did not move into each other, so the engine computed the contact wrong. Second, reruns from one step before the artifact show that clearing the contact cache or reducing the contact offset removes the false contact. Third, PhysX's contact code, run outside the simulator on the recorded poses, returns the same false contact, which we trace through the code.

\paragraph{The reported contact.}
In the episode of \cref{fig:ejection}, the 27.5\,mm overlap at onset comes from a convex piece of the can and a 12 by 20 by 15\,mm piece of the basket wall. Their hulls are 21\,mm apart when the step begins, beyond the two 10\,mm contact offsets, and the pair is absent from the contact report one step earlier. In 35 of the 37 Place Can Basket episodes with an artifact, the deepest can--basket contact at the first onset is deeper than 5\,mm. Its two hulls are then a median 15.0\,mm apart at the start of the step, more than 2\,mm in 33 and more than 20\,mm in 7. PhysX's normal points from the second shape to the first. In all 1958 can--basket contacts one step before onset, the can's piece lies on that side; in 34 of the 35 onset contacts it lies on the other side. The reported separation equals the gap plus the thickness of the piece, measured along the normal from the can vertex farthest from the basket; the median difference is below 0.01\,mm. One step earlier, the same pair reports a positive separation of 0.5 to 20\,mm in 21 episodes and nothing in the others.

\paragraph{The path in PhysX~5.3.1.}
PhysX computes contacts between two convex pieces from their closest points, found by GJK~\citep{gilbert1988gjk}. Each GJK iteration adds a support point to a simplex and reduces it to the feature closest to the origin. PhysX~5.3.1 treats a triangle $abc$ as degenerate, and falls back to a segment, when $|ab \times ac|^2 \le \epsilon$. Here $\epsilon = 2^{-23} \approx 1.19 \times 10^{-7}$ is the single precision machine epsilon, with lengths in meters. The test does not scale with the geometry: a right triangle with 1.8\,cm legs already fails it. At the onset steps of four Place Can Basket episodes, the triangles have $|ab \times ac|^2$ of $8.81 \times 10^{-8}$, $1.189 \times 10^{-7}$, $1.180 \times 10^{-7}$ and $5.99 \times 10^{-8}$, with edges of 11 to 96\,mm. The triangle is rejected, the distance stops decreasing, and GJK leaves through its exit for a degenerate simplex. That exit returns points that are not the closest and skips the check against the contact distance. In the episode of \cref{fig:ejection} it returns 26.95\,mm for pieces 21.1\,mm apart. Contact generation accepts this result when the search direction and the normal agree, and builds contact points from the faces that hold the two closest points. The point on one piece lies inside it or on its far side, so the selected face, which sets the normal, points away from the other piece. Measured to that face's plane, each separation is negative and equals the gap plus the thickness of the piece. Points are reported whenever their separation is below the contact offset, with no lower bound, and the solver removes the depth within one step. PhysX~5.4.0 uses $\epsilon^2 \approx 1.42 \times 10^{-14}$. This is the only change to the simplex reduction between the two versions, and the later steps are unchanged up to 5.9.0.

\paragraph{Reproduction outside the simulator.}
A short program builds the two pieces from their vertices, keeps its own contact cache, and calls PhysX's contact generation on the recorded poses from the step the pieces first come within range. For four Place Can Basket episodes it returns the simulator's onset normal and separations to the printed digits (\cref{tab:versions}). Taking only the header of the simplex reduction from 5.4.0 removes all four contacts. Taking only the other 5.4.0 change to this routine, a corrected extent of the second hull, keeps all four. In the episode of \cref{fig:ejection}, the false contact needs the simplex that GJK cached five steps earlier; in two of the four it also appears from an empty cache. In the simulator, writing the can's pose back unchanged just before onset clears its cached pairs and removes the artifact in 6 of 9 episodes.

On Stack Bowls Three we traced the 21 bowl--bowl onset contacts deeper than 5\,mm in the 12 episodes with an artifact, using a separating axis test for the true distance. In 16, the pieces do not intersect, with gaps of 0 to 21.7\,mm, and the reported depth of 6 to 58\,mm comes from the path above. The wrong face is on the first piece in 4 and on the second in 7. In the other 5 the reference face is right, but its partner face is nearly perpendicular to the normal. In 4, all late in one episode, the pieces do intersect, and the expanding polytope algorithm returns the true depth of 11.7 to 18.0\,mm. The program does not reproduce the last one. With the 5.4.0 threshold, none of the 16 is reported and the 4 real ones keep their depth.

\begin{tableH}
  \centering
  \caption{\textbf{The false contact across PhysX versions.} Onset contacts returned outside the simulator for the recorded poses. A dash: not run.}
  \label{tab:versions}
  \small
  \begin{tabular}{lccc}
    \toprule
    Version & Can, false (4) & Bowls, false (16) & Bowls, real (4) \\
    \midrule
    5.3.1 & 4 & 16 & 4 \\
    5.3.1 with the simplex reduction of 5.4.0 & 0 & 0 & 4 \\
    5.3.1 with the hull extent of 5.4.0 & 4 & -- & -- \\
    5.4.0, 5.4.2, 5.5.1, 5.6.1 & 0 & -- & -- \\
    5.9.0 & 0 & 0 & 4 \\
    \bottomrule
  \end{tabular}
\end{tableH}

\paragraph{Inside the simulator.}
SAPIEN bundles PhysX in its own library, and releases 3.0.0 and 3.0.3 still contain 5.3.1. In a copy of the SAPIEN~3.0.0b1 library we change only the constant that the four triangle tests of the simplex reduction load, from $2^{-23}$ to $2^{-46}$. The unchanged copy replays the episode of \cref{fig:ejection} bit for bit over 60922 steps. In the repaired copy, the recorded actions of four Place Can Basket episodes give no can--basket contact deeper than 5\,mm and a peak can speed below 1\,m/s, against 27.5\,mm and 21.8\,m/s in that episode. These replays leave the recording between steps 3499 and 5322, long before the artifact near step 11000, because the threshold also affects regular contacts during the grasp. We therefore evaluate the repaired engine with the scripted expert (\cref{tab:threshold}) and in closed loop (\cref{tab:repaired}) rather than by paired replay. For the expert we run RoboTwin's demonstration on seeds 100000 to 100059 in both engines; below, the released engine comes first. On Place Can Basket the plan succeeds on 57 and 60 seeds and the demonstration on 51 and 59; on Stack Bowls Three, on 35 and 36, and on 30 and 36. On seeds where both engines finish, 8 and 6 succeed only in the repaired engine, and 0 and 1 only in the released one.

\paragraph{Contact offset.}
A shape makes contacts with every shape closer than the sum of their contact offsets, 10\,mm each in SAPIEN. When we change the offset of the task's objects and containers 0.1\,s before the first artifact, fewer Place Can Basket episodes keep an artifact as the offset shrinks (\cref{tab:offset}). All 37 replays match their recordings up to the switch. On Stack Bowls Three at 1\,mm, 6 of the 12 episodes keep an artifact and 4 failures pass, since nested bowls have pieces closer than any offset.

\begin{tableH}
  \centering
  \caption{\textbf{Contact offset} of can and basket, changed 0.1\,s before the first artifact, in the 37 Place Can Basket episodes of LingBot-VA with an artifact: 152 events, 21 failures, 16 successes. The last row is the recording.}
  \label{tab:offset}
  \small
  \begin{tabular}{lcccc}
    \toprule
    Contact offset & Episodes with an artifact & Events & Failures that pass & Successes kept \\
    \midrule
    0.5\,mm & 3  & 10  & 18/21 & 10/16 \\
    1\,mm   & 4  & 14  & 19/21 & 12/16 \\
    2\,mm   & 9  & 29  & 18/21 & 14/16 \\
    3\,mm   & 11 & 25  & 18/21 & 12/16 \\
    5\,mm   & 13 & 46  & 16/21 & 10/16 \\
    10\,mm  & 37 & 152 & --    & -- \\
    \bottomrule
  \end{tabular}
\end{tableH}

\begin{tableH}
  \centering
  \caption{\textbf{One constant.} \emph{False}: the 20 traced onset contacts whose pieces do not intersect, recomputed outside the simulator. \emph{Real}: 4 onset contacts whose pieces intersect. \emph{Expert}: RoboTwin's scripted demonstration, 60 seeds per task; episodes with an artifact, and episodes that fail after a valid plan.}
  \label{tab:threshold}
  \small
  \begin{tabular}{@{}lcccccc@{}}
    \toprule
    & \multicolumn{2}{c}{Contacts} & \multicolumn{2}{c}{Expert, artifact} & \multicolumn{2}{c}{Expert, failed} \\
    \cmidrule(lr){2-3}\cmidrule(lr){4-5}\cmidrule(l){6-7}
    Engine & False (20) & Real (4) & Can & Bowls & Can & Bowls \\
    \midrule
    PhysX~5.3.1 (RoboTwin)                & 20 & 4 & 18 & 6 & 6 & 5 \\
    PhysX~5.3.1, threshold of 5.4.0       & 0  & 4 & 1  & 0 & 1 & 0 \\
    PhysX~5.9.0                           & 0  & 4 & -- & -- & -- & -- \\
    \bottomrule
  \end{tabular}
\end{tableH}

\paragraph{Closed loop evaluation in the repaired engine.}
\Cref{tab:repaired,tab:closed-loop-arms} evaluate LingBot-VA, with its sampling noise reseeded at every reset, on 20 scored seeds per task in four arms: RoboTwin as released; a 1\,mm contact offset on objects and containers, set at the policy's first action so that the scripted demonstration keeps the released settings; the repaired engine, for both policy and demonstration; and the persistent contact manifold off for the whole scene. The scripted demonstration is not deterministic, so the arms score slightly different seeds, and paired counts use the seeds scored in both. On the five container tasks, the repaired engine shares 83 seeds with the released one: 73 succeed in both, 8 only in the repaired engine and 2 only in the released one. The offset arm shares 85 seeds: 74, 8 and 2, and one fails in both. With the manifold off, 3 episodes still contain an artifact, 2 of them failures; the repaired engine has none. We compare HyVLA in the repaired engine with its audited run, which was not reseeded. They share 83 scored seeds; 7 succeed only in the repaired engine and 4 only in the audited run. Its one artifact in the repaired engine is a bread thrown off the other bread at 2.4\,m/s on Place Bread Basket. The scripted demonstration on that seed shows the same event. On 40 more seeds per task, run in both engines, HyVLA succeeds in 179 of 200 episodes as released, 30 of them with an artifact, and in 196 of 200 in the repaired engine, none with an artifact.

\begin{tableH}
  \centering
  \caption{\textbf{Closed loop evaluation, all arms.} LingBot-VA, 20 scored seeds per task, noise reseeded every episode. Cells: successes, episodes with an artifact after the gravity stage in parentheses, and seeds where the scripted demonstration failed. \emph{Manifold off}: persistent contact manifold off for the whole scene.}
  \label{tab:closed-loop-arms}
  \small
  \begin{tabular}{lcccc}
    \toprule
    Task & Benchmark & Offset 1\,mm & Repaired & Manifold off \\
    \midrule
    Place Can Basket    & 15 (5), 4  & 19 (3), 4  & 20 (0), 2  & 20 (0), 3 \\
    Stack Bowls Three   & 15 (6), 18 & 17 (3), 16 & 19 (0), 11 & 18 (2), 12 \\
    Place Object Basket & 20 (0), 14 & 20 (0), 12 & 20 (0), 12 & 20 (1), 3 \\
    Place Bread Basket  & 19 (1), 6  & 19 (0), 5  & 20 (0), 4  & 20 (0), 3 \\
    Stack Bowls Two     & 20 (0), 7  & 20 (0), 7  & 19 (0), 7  & 19 (0), 7 \\
    Blocks Ranking Size & 18 (0), 1  & 17 (0), 1  & 19 (0), 2  & 17 (0), 3 \\
    Open Microwave      & 15 (0), 20 & 18 (0), 20 & 16 (0), 20 & 17 (0), 37 \\
    \midrule
    Five container tasks & 89 (12) & 95 (6) & 98 (0) & 97 (3) \\
    \bottomrule
  \end{tabular}
\end{tableH}

\subsection{Stress test}
\label{app:stress}
The scene holds RoboTwin's can, model~1 of object 071 with 22 convex pieces and 10\,g, and basket, model~1 of object 110 with 54 convex pieces and 0.5\,kg, on a table. SAPIEN~3.0.0b1 with PhysX and MuJoCo~3.3.1 share the collision pieces, masses, inertias and center of mass frames, with gravity on and a 4\,ms step. The basket takes its pose from the episode of \cref{fig:ejection} one step before the ejection. Each of 300 can poses has a uniform random orientation and a position within 3\,cm of the recorded one horizontally and from 3\,cm below to 1\,cm above it. The can starts at rest, 288 poses start at least 5\,mm deep in the basket, and each pose runs for 0.5\,s (\cref{tab:stress}).

\begin{tableH}
  \centering
  \caption{\textbf{Stress test.} RoboTwin's can at 300 resting poses inside its basket, in PhysX and in MuJoCo with the same assets and step. \emph{Thrown}: at least 1\,m/s and 4\,cm away. \emph{Removal}: initial overlap over exit speed. RoboTwin's values in parentheses.}
  \label{tab:stress}
  \small
  \begin{tabular}{lcccc}
    \toprule
    Engine and setting & Thrown (\%) & Exit speed (m/s) & Distance (cm) & Removal (ms) \\
    \midrule
    PhysX, RoboTwin settings           & 91 & 8.77  & 155 & 2.6 \\
    PhysX, can mass 100\,g (10\,g)     & 90 & 6.86  & 73  & 3.3 \\
    PhysX, solver iterations 32/8 (10/1) & 90 & 16.78 & 200 & 1.4 \\
    PhysX, depenetration $\le$ 1\,m/s  & 65 & 1.36  & 10  & 16.5 \\
    PhysX, depenetration $\le$ 0.1\,m/s & 0 & 0.14  & 3.7 & 169 \\
    MuJoCo, default (time constant 20\,ms) & 5 & 0.56 & 2.8 & 38.6 \\
    MuJoCo, time constant 8\,ms        & 42 & 1.99  & 3.8 & 10.6 \\
    \bottomrule
  \end{tabular}
\end{tableH}

Mass and solver iterations leave the thrown share at 90\%. Only the maximum depenetration velocity changes it, the parameter that PhysX's documentation names for bodies that start in overlap~\citep{physx54}.

The response depends on depth. Pushed straight down into the basket floor, the can leaves PhysX with RoboTwin's settings at a peak of 0.04\,m/s within 0.1\,s from depths up to 10\,mm, and 4.20\,m/s from 20\,mm. From 20\,mm, depenetration limits of 1 and 0.1\,m/s give 1.24 and 0.14\,m/s, and 32 position iterations give 11.92\,m/s. MuJoCo's default contact gives 0.24\,m/s from 10\,mm and 1.05\,m/s from 20\,mm. Bullet's default contact gives 0.26 and 1.02\,m/s, and 2.63 and 10.22\,m/s with its error reduction parameter raised to 0.8; Bullet's shapes carry a 1\,mm margin, so each of its overlaps is 2\,mm deeper. MuJoCo rejects time constants below two physics steps; with this check off, 4 or 1\,ms make the scene diverge. From the recorded poses one step before the ejection, the can does not overlap the basket in this scene and leaves at 0.23\,m/s, as \cref{app:cause} predicts. The test therefore starts from real overlaps of the onset depth.

\begin{figureH}
  \centering
  \includegraphics[width=\linewidth]{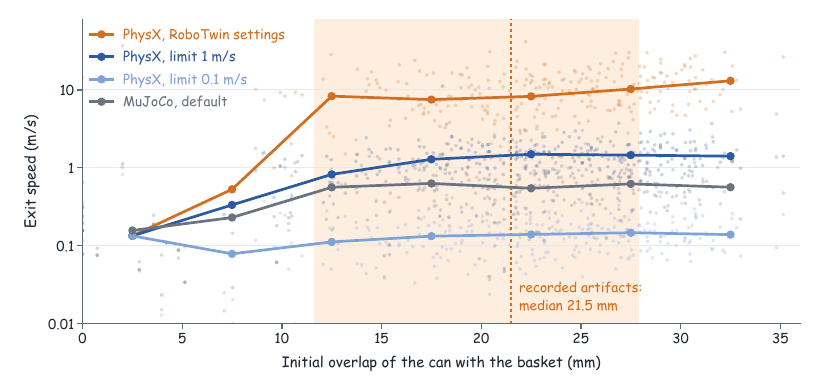}
  \caption{\textbf{Exit speed against initial overlap} in the stress test of \cref{tab:stress}: PhysX with RoboTwin's settings and with depenetration limits of 1 and 0.1\,m/s, and MuJoCo's default contact. Dots: poses; lines: medians in 5\,mm bins. With RoboTwin's settings the median exit speed is 0.13\,m/s below 10\,mm of overlap and 9.4\,m/s at or above it. The limits and MuJoCo stay between 0.14 and 1.5\,m/s at every depth. The band marks the onset overlap of the 99 recorded Place Can Basket artifacts, median 21.5\,mm and interquartile range 11.6 to 27.9\,mm, above the depth where PhysX starts to eject.}
  \label{fig:depth}
\end{figureH}

\subsection{Rigid IPC unit tests}
\label{app:rigidipc}
Rigid IPC~\citep{ferguson2021rigidipc} guarantees that no step ends in an intersection, and its repository provides unit tests of hard contact cases, five of them from the test set of Erleben. We run 13 of these scenes for 5\,s each in three simulators: Rigid IPC itself, MuJoCo and PhysX. Rigid IPC runs its own scene files with their 10\,ms step. MuJoCo and PhysX run the scenes that the repository provides for MuJoCo, with the same 2\,ms step, the convex hull of each mesh, zero friction and the same density; the flat ground mesh becomes a slab 0.1\,m thick under the same surface. \sg{} uses the thresholds of \cref{tab:detector}, every free body as a target and every fixed body as a container, and events pass the same gravity filter as on RoboTwin. It is attached to MuJoCo and PhysX and reads the recorded trajectory of Rigid IPC. Rigid IPC records the smallest distance in the scene and no contact pairs, so the detector receives a contact between every pair of bodies whenever that distance is below 5\,mm, which can only add flags.

These scenes have no robot, so no body can move much faster than free fall. For each scene we take the largest ratio of a body's speed to the free fall speed from the highest starting point in the scene, and count an engine error when the ratio exceeds 1.5 (\cref{tab:rigidipc}). In Rigid IPC, no scene has an error, the smallest distance between two bodies stays above 0.09\,mm, and \sg{} flags no scene. MuJoCo shows no error either, and \sg{} flags no scene. PhysX shows one, on large mass ratio: a heavy cube lands on a light one, their overlap reaches 32\,mm, and the light cube leaves at 63.24\,m/s, 4.73 times the free fall speed. In Rigid IPC the same two cubes come to rest, and no body exceeds 3.04\,m/s. \sg{} flags this PhysX scene and no other. The smaller threshold of \cref{app:cause} leaves the PhysX numbers unchanged: here the overlap is real, and the engine removes it in one step.

\begin{tableH}
  \centering
  \caption{\textbf{Rigid IPC unit tests.} \emph{Peak}: peak speed of a free body in m/s. \emph{Ratio}: largest ratio of a body's speed to the free fall speed from the highest starting point. \emph{Flags}: events of \sg{} that pass the gravity filter. In several scenes the body slides off its support and keeps falling in all three simulators, which gives peaks above 40\,m/s at a ratio of 1.}
  \label{tab:rigidipc}
  \small
  \setlength{\tabcolsep}{4pt}
  \begin{tabular}{lccccccccc}
    \toprule
    & \multicolumn{3}{c}{Rigid IPC, 10\,ms} & \multicolumn{3}{c}{MuJoCo, 2\,ms} & \multicolumn{3}{c}{PhysX, 2\,ms} \\
    \cmidrule(lr){2-4}\cmidrule(lr){5-7}\cmidrule(lr){8-10}
    Scene & Peak & Ratio & Flags & Peak & Ratio & Flags & Peak & Ratio & Flags \\
    \midrule
    5 cubes & 10.29 & 0.99 & 0 & 10.38 & 1.00 & 0 & 10.24 & 1.00 & 0 \\
    edge edge parallel & 44.58 & 1.00 & 0 & 44.00 & 1.00 & 0 & 44.09 & 1.00 & 0 \\
    edge edge & 18.34 & 0.98 & 0 & 2.39 & 1.00 & 0 & 2.37 & 1.00 & 0 \\
    Erleben, cliff edges & 0.88 & 0.95 & 0 & 0.98 & 0.99 & 0 & 0.96 & 0.99 & 0 \\
    Erleben, internal edges & 0.88 & 0.95 & 0 & 0.98 & 0.99 & 0 & 0.96 & 0.99 & 0 \\
    Erleben, spike and wedge & 47.51 & 1.00 & 0 & 43.68 & 1.00 & 0 & 43.62 & 1.00 & 0 \\
    Erleben, spikes & 46.63 & 1.00 & 0 & 43.67 & 1.00 & 0 & 43.62 & 1.00 & 0 \\
    Erleben, wedges & 47.88 & 1.00 & 0 & 45.54 & 1.00 & 0 & 45.51 & 1.00 & 0 \\
    face vertex & 44.25 & 1.00 & 0 & 44.02 & 1.00 & 0 & 44.09 & 1.00 & 0 \\
    large mass ratio & 3.04 & 0.98 & 0 & 3.14 & 1.00 & 0 & 63.24 & 4.73 & 4 \\
    tet corner & 4.51 & 0.99 & 0 & 4.65 & 1.00 & 0 & 3.00 & 0.99 & 0 \\
    vertex face & 2.25 & 0.98 & 0 & 2.39 & 1.00 & 0 & 2.37 & 1.00 & 0 \\
    vertex vertex & 3.44 & 0.97 & 0 & 20.97 & 1.00 & 0 & 14.72 & 1.00 & 0 \\
    \bottomrule
  \end{tabular}
\end{tableH}

\subsection{Reruns}
\label{app:reruns}
\begin{figureH}
  \centering
  \includegraphics[width=\linewidth]{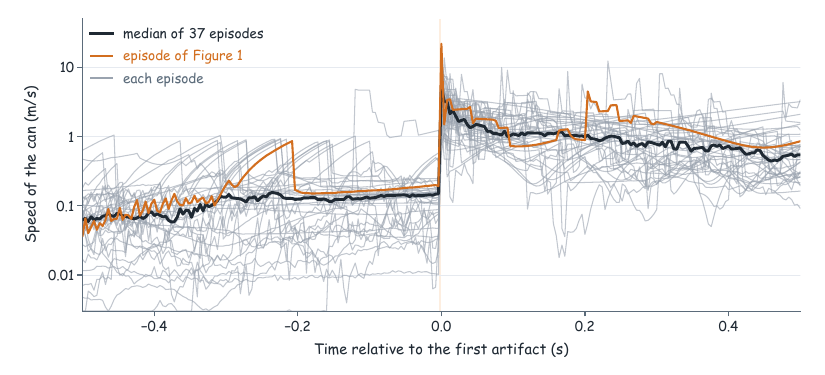}
  \caption{\textbf{The can's speed around the first artifact} in the 37 Place Can Basket episodes of LingBot-VA with one (21 failures, 16 successes), at every 4\,ms step from 0.5\,s before to 0.5\,s after onset. Gray: episodes; bold: median; orange: the episode of \cref{fig:ejection}. One step before onset the median speed is 0.15\,m/s. The largest one step rise falls on the onset step in 33 episodes and within four steps in the other 4. It ranges from 0.53 to 21.6\,m/s, with a median of 5.7\,m/s.}
  \label{fig:traces}
\end{figureH}

\paragraph{Contact manifold.}
\label{app:pcm}
PhysX keeps the warm start of GJK in its persistent contact manifold, which RoboTwin leaves on. With only the manifold switched off, PhysX uses another contact routine. Replayed from the first step, the recorded actions then give an artifact in 2 of the 37 Place Can Basket episodes and in 1 of the 8 Stack Bowls Three episodes. Events fall from 152 to 3 and from 83 to 1, and the median peak speed from 7.89 to 0.91\,m/s and from 5.22 to 1.31\,m/s. The success check passes 27 and 6 of these replays, against 16 and 0 as recorded. The replays diverge from the recording at the first contact, so they are not paired with it.

\paragraph{Changes from the first physics step.}
Applied from the first step, every change of a simulation parameter also rescues Place Can Basket failures without an artifact (\cref{tab:first-step}), so these replays cannot tell the parameters apart. A clamp that caps the object's speed at 2\,m/s after each step is not a simulation parameter, and it rescues none of them.

\begin{tableH}
  \centering
  \caption{\textbf{Replays from the first physics step} of the recorded failures: failures that pass the success check.}
  \label{tab:first-step}
  \small
  \begin{tabular}{lccc}
    \toprule
    & \multicolumn{2}{c}{Place Can Basket failures} & Stack Bowls Three \\
    \cmidrule(lr){2-3}
    Change & with artifact (21) & without artifact (2) & failures (12) \\
    \midrule
    Depenetration $\le$ 1\,m/s   & 19 & 2 & 8 \\
    Depenetration $\le$ 0.1\,m/s & 9  & 2 & 5 \\
    Object mass 100\,g           & 17 & 2 & 6 \\
    Solver iterations 32/8       & 14 & 1 & 5 \\
    Speed clamp 2\,m/s           & 14 & 0 & 0 \\
    \bottomrule
  \end{tabular}
\end{tableH}

\paragraph{One setting changed before the artifact.}
Switched on 0.1\,s before the first artifact, a smaller contact offset and the depenetration limit rescue the most failures (\cref{tab:knobs}). A ten times heavier object and more solver iterations rescue fewer and leave the object faster.

\begin{tableH}
  \centering
  \caption{\textbf{One setting switched on 0.1\,s before the first artifact,} LingBot-VA episodes with artifacts. \emph{Rescued}: failures that pass. \emph{Peak}: median object speed after the switch over all these episodes. Offset changed on the task's objects and containers; other settings as in \cref{sec:method-rerun}.}
  \label{tab:knobs}
  \small
  \begin{tabular}{lcccc}
    \toprule
    & \multicolumn{2}{c}{Place Can Basket} & \multicolumn{2}{c}{Stack Bowls Three} \\
    \cmidrule(lr){2-3}\cmidrule(l){4-5}
    Change & Rescued & Peak (m/s) & Rescued & Peak (m/s) \\
    \midrule
    Contact offset 1\,mm (10\,mm)   & 19/21 & 0.91  & 4/12 & 2.50 \\
    Depenetration $\le$ 1\,m/s      & 18/21 & 1.39  & 7/12 & 0.65 \\
    Object mass 100\,g              & 14/21 & 3.72  & 5/12 & 2.12 \\
    Solver iterations 32/8          & 12/21 & 11.20 & 4/12 & 7.00 \\
    \bottomrule
  \end{tabular}
\end{tableH}

\paragraph{Limit value.}
Switched on 0.1\,s before the first artifact, limits from 0.1 to 1\,m/s rescue nearly the same failures; 2\,m/s rescues fewer, and the object leaves faster (\cref{tab:limit-sweep}). Peak speeds here are medians over the 21 failures; \cref{tab:knobs} also counts the 16 successes, hence 1.39 against 1.25\,m/s. All 108 replays of this sweep match their recordings up to the switch.

\begin{tableH}
  \centering
  \caption{\textbf{Limit value.} LingBot-VA failures with artifacts that pass with the limit switched on 0.1\,s before the first artifact, and their median peak object speed after the switch. \emph{Three tasks}: Place Object Basket, Place Bread Basket and Stack Bowls Two, one failure each. Successes with an artifact were replayed at 1\,m/s only.}
  \label{tab:limit-sweep}
  \small
  \begin{tabular}{lcccccc}
    \toprule
    & \multicolumn{2}{c}{Place Can Basket (21)} & \multicolumn{2}{c}{Stack Bowls Three (12)} & Three tasks (3) & Five tasks \\
    \cmidrule(lr){2-3}\cmidrule(lr){4-5}\cmidrule(lr){6-6}\cmidrule(l){7-7}
    Limit & Rescued & Peak (m/s) & Rescued & Peak (m/s) & Rescued & Rescued \\
    \midrule
    0.1\,m/s & 18 & 0.91 & 7 & 0.40 & 3 & 28/36 \\
    0.5\,m/s & 18 & 1.05 & 7 & 0.63 & 2 & 27/36 \\
    1\,m/s   & 18 & 1.25 & 7 & 0.65 & 3 & 28/36 \\
    2\,m/s   & 16 & 1.95 & 4 & 1.09 & 3 & 23/36 \\
    \bottomrule
  \end{tabular}
\end{tableH}

\paragraph{Successes with an artifact.}
We run the same replay on every success with an artifact. \Cref{tab:counts} lists how many stay successes, per task and policy.

\paragraph{Closed loop evaluation with the limit on.}
We also rerun LingBot-VA with the 1\,m/s depenetration limit on for the whole policy rollout, keeping the scripted demonstration unchanged (\cref{tab:closed-loop-depen}). Place Can Basket rises from 67 to 81 of 90 episodes over three repetitions, and both bowl tasks reach 20 of 20. Place Bread Basket and Place Object Basket fall, the latter with an artifact still in 1 of 20 episodes. Tasks without container artifacts move both ways. The policy samples its actions without a fixed seed, so each closed loop episode is a new rollout. The as-run rate on the shared seeds counts each seed once, with its recorded outcome. Put Bottles Dustbin was not completed. HyVLA, run the same way on the five container tasks, keeps 16 of 20 on Place Can Basket, rises on both bowl tasks and falls on the other two. On Place Object Basket 4 of its 18 seeds ran twice, and 7 of its 102 episodes still contain an artifact.

\begin{tableH}
  \centering
  \caption{\textbf{Closed loop evaluation with the limit on.} Successes as reported and with the depenetration limit of 1\,m/s on for the whole policy rollout. $^\ast$The first 20 scored seeds of the reported run of 50.}
  \label{tab:closed-loop-depen}
  \small
  \begin{tabular}{lcc}
    \toprule
    Task & Reported & Limit on \\
    \midrule
    \multicolumn{3}{@{}l}{\emph{LingBot-VA}} \\
    Place Can Basket    & 67/90 & 81/90 \\
    Stack Bowls Three   & 16/20$^\ast$ & 20/20 \\
    Stack Bowls Two     & 19/20 & 20/20 \\
    Place Bread Basket  & 19/20 & 17/20 \\
    Place Object Basket & 18/20 & 10/20 \\
    Stamp Seal          & 19/20 & 19/20 \\
    Blocks Ranking Size & 17/20 & 18/20 \\
    Open Microwave      & 15/20 & 10/20 \\
    Turn Switch         & 9/20  & 9/20 \\
    Put Object Cabinet  & 7/10  & 5/10 \\
    Place Burger Fries  & 20/20 & 20/20 \\
    \midrule
    \multicolumn{3}{@{}l}{\emph{HyVLA}} \\
    Place Can Basket    & 16/20 & 16/20 \\
    Stack Bowls Three   & 18/20 & 20/20 \\
    Stack Bowls Two     & 17/20 & 20/20 \\
    Place Bread Basket  & 20/20 & 14/20 \\
    Place Object Basket & 19/20 & 10/22 \\
    \bottomrule
  \end{tabular}
\end{tableH}

\paragraph{Closed loop evaluation with changed physics.}
Changing one parameter for the whole closed loop evaluation also changes the reported rate (\cref{tab:closed-loop-reps}). A 100\,g can raises LingBot-VA on Place Can Basket from 74.4\% with RoboTwin's 10\,g can to 95.6\%. A paired test over seeds gives $p = 0.0015$, and the 95\% interval of the difference is $[+10.3, +32.8]$ points. More solver iterations give 86.7\%, with $p = 0.17$ under a Wilcoxon signed rank test on success per seed.

\begin{tableH}
  \centering
  \caption{\textbf{Closed loop success} of LingBot-VA on Place Can Basket under three physics configurations, 30 scored seeds per repetition. The solver configuration ran two repetitions.}
  \label{tab:closed-loop-reps}
  \small
  \begin{tabular}{lccccc}
    \toprule
    Configuration & Rep.~1 & Rep.~2 & Rep.~3 & Total & Mean $\pm$ std (\%) \\
    \midrule
    RoboTwin (can 10\,g) & 23/30 & 24/30 & 20/30 & 67/90 & $74.4 \pm 6.9$ \\
    Can mass 100\,g & 29/30 & 30/30 & 27/30 & 86/90 & $95.6 \pm 5.1$ \\
    Solver position iterations 32 & 26/30 & 26/30 & -- & 52/60 & $86.7 \pm 0.0$ \\
    \bottomrule
  \end{tabular}
\end{tableH}

\paragraph{Closed loop takeover.}
We also let the policy take over from a replayed prefix 1\,s before the first artifact of each of the 21 Place Can Basket failures, with the speed clamp on or the physics unchanged. All 42 prefixes match their recordings bit for bit. Unchanged physics rescues 10 failures and the clamp 15; 7 are rescued only with the clamp and 2 only without, with $p = 0.18$ under a McNemar exact test. From the same state the policy's own sampling often avoids the artifact, so the corrected rate of \cref{eq:corrected} replays the recorded actions instead.

\paragraph{Skipped scripted demonstrations.}
RoboTwin scores a policy on a seed only if its scripted demonstration passes there first. On Place Can Basket, 22 seeds were skipped this way over the three repetitions, and the demonstration on 16 of them has a contact artifact. With the depenetration limit switched on 0.1\,s before the first artifact, 13 of the 16 pass, and all 16 prefixes match the recording up to the switch (\cref{fig:trajectories}, bottom). The artifacts thus also decide which seeds enter the evaluation, before any policy is scored (\cref{tab:expert}).

\subsection{Short limit on one body, and a control}
\label{app:narrow}
The replay above keeps the limit on every body until the end, so a flip may come from any later contact. We fixed the following plan before running it. Let $t_0$ be the onset of the first artifact and the anchor its ejected body. \emph{Body} limits the anchor only, from $t_0 - 0.1$\,s to $t_0 + 0.2$\,s, then restores its value. \emph{Partner} also limits the body the anchor touched just before $t_0$. \emph{Resolved} limits the anchor only until its overlap stays within its rest offset plus 1\,mm for five 4\,ms steps, at most 1\,s after $t_0$ and a median of 6 steps. Each form runs at limits of 0.5, 1 and 2\,m/s on the flagged episodes: 59 of LingBot-VA and 10 of HyVLA on the five tasks, and 8 more of LingBot-VA on other tasks. As a control, 113 LingBot-VA episodes, 97 on the five tasks, and 20 HyVLA episodes without an artifact, matched on task and outcome, get each form when the gripper first touches an object, with that object as the anchor. All 2{,}121 replays match their recordings bit for bit up to the switch. For LingBot-VA every form keeps most of the rise from 83.1\%; for HyVLA the short forms give 90.0 to 93.0\%, against 90.0\% reported. No form changes a control episode: at 1\,m/s, 0 of 106 LingBot-VA successes fail, with a 95\% upper bound of 3.5\%, and 0 of 7 failures pass; for HyVLA, 0 of 17 and 0 of 3.

\begin{tableH}
  \centering
  \caption{\textbf{Short limits and control,} five RoboTwin tasks. \emph{Flipped}: failures that pass and successes that fail among the flagged episodes, LingBot-VA 40 and 19, HyVLA 6 and 4. \emph{Rate}: \cref{eq:corrected} at each limit. \emph{Control}: episodes without an artifact on the five tasks, LingBot-VA 97 and HyVLA 20, with the same forms at the first gripper contact.}
  \label{tab:narrow}
  \small
  \begin{tabular}{@{}lccccc@{}}
    \toprule
    & \multicolumn{2}{c}{Flipped at 1\,m/s} & \multicolumn{3}{c}{Rate (\%) at limit} \\
    \cmidrule(lr){2-3}\cmidrule(l){4-6}
    Form & F$\to$S & S$\to$F & 0.5 & 1 & 2 \\
    \midrule
    \multicolumn{6}{@{}l}{\emph{LingBot-VA, reported 83.1\%}} \\
    Every body, to the end & 32/40 & 9/19  & 91.5 & 91.9 & 89.6 \\
    Body                   & 21/40 & 10/19 & 87.7 & 87.3 & 86.2 \\
    Partner                & 23/40 & 10/19 & 88.8 & 88.1 & 86.9 \\
    Resolved               & 19/40 & 11/19 & 87.3 & 86.2 & 86.2 \\
    \midrule
    Control, any short form & 0/4 & 0/93 & 83.1 & 83.1 & 83.1 \\
    \midrule
    \multicolumn{6}{@{}l}{\emph{HyVLA, reported 90.0\%}} \\
    Every body, to the end & 5/6 & 2/4 & 94.0 & 93.0 & 90.0 \\
    Body                   & 4/6 & 3/4 & 93.0 & 91.0 & 91.0 \\
    Partner                & 4/6 & 4/4 & 93.0 & 90.0 & 91.0 \\
    Resolved               & 4/6 & 1/4 & 93.0 & 93.0 & 91.0 \\
    Control, any short form & 0/3 & 0/17 & 90.0 & 90.0 & 90.0 \\
    \bottomrule
  \end{tabular}
\end{tableH}

\paragraph{Which score to use.} The four scores of \cref{tab:scores} are references. We use the replay rate as the point estimate because it keeps the recorded actions and changes the physics of one body for 0.3\,s around the artifact; it measures how sensitive the recorded execution is to that change.

\subsection{Five models against a real robot}
\label{app:rank}
We compare the ranking of five models in RoboTwin as released, in the repaired engine and on a real robot. In simulation, each model runs the five container tasks in both engines, 20 episodes per task; the rows of LingBot-VA and HyVLA are those of \cref{tab:repaired}. On the robot, a dual-arm Piper-X, we fine-tune each model from its released weights on 100 real demonstrations of each task and test it 10 times on each of three tasks: stacking bowls, stacking blocks and placing objects into a basket.

As released, RoboTwin ranks HyVLA first, although it is fourth on the robot (\cref{tab:rank}). In the repaired engine, the five models fall into the same order as on the robot. Spearman's correlation with the real ranking is 0.40 as released and 0.97 in the repaired engine.

\begin{tableH}
  \centering
  \caption{\textbf{Five models in simulation and on a real robot.} Successes over five tasks in RoboTwin and three tasks on the robot.}
  \label{tab:rank}
  \small
  \begin{tabular}{lccc}
    \toprule
    Model & Released & Repaired & Real robot \\
    \midrule
    LingBot-VA   & 89/100 & 98/100 & 24/30 \\
    InternVLA-A1 & 86/100 & 98/100 & 23/30 \\
    HyVLA        & 90/100 & 94/100 & 20/30 \\
    LingBot-VLA  & 82/100 & 95/100 & 21/30 \\
    StarVLA      & 60/100 & 71/100 & 10/30 \\
    \bottomrule
  \end{tabular}
\end{tableH}

\section{Other Platforms}
\label{app:platforms}

This section audits RoboLab, RoboDojo, LIBERO and RoboCasa; \cref{app:impl} describes the adapters. \Cref{tab:all-bench} gives every success rate before and after removing artifacts. A rate changes only when an artifact decides an episode that would otherwise succeed, so it moves little for weak policies. An artifact can also occur in a success; removing that episode lowers the rate.

\subsection{RoboLab}
\label{app:robolab}

\paragraph{Setup.}
NVIDIA RoboLab\footnote{\url{https://github.com/NVlabs/RoboLab}} runs Isaac Sim~5.0 and Isaac Lab~2.2 with GPU PhysX. We run $\pi_{0.5}$-DROID~\citep{intelligence2025pi05} and GR00T~N1.7-DROID~\citep{bjorck2025gr00t}. Each policy runs all 120 tasks with the default instruction, 20 parallel episodes per task: 2,400 episodes per policy, 4,800 in total. The detector runs online with the parameters of \cref{tab:detector}, fixed before the run.

\begin{tableH}
  \centering
  \caption{\textbf{RoboLab scores} in \%, 2,400 episodes per policy, with 674 and 212 successes. \emph{All flags}: every flagged episode counts as an artifact. \emph{Reviewed}: precision and miss rate from the human review (\cref{app:accuracy}). Interval: lowest and highest score if each artifact episode could have gone either way.}
  \label{tab:robolab}
  \small
  \begin{tabular}{@{}lcccccc@{}}
    \toprule
    & & & \multicolumn{2}{c}{All flags} & \multicolumn{2}{c}{Reviewed} \\
    \cmidrule(lr){4-5}\cmidrule(l){6-7}
    Policy & Flagged & Reported & Exclusion & Interval & Exclusion & Interval \\
    \midrule
    $\pi_{0.5}$-DROID & 496 (20.7) & 28.1 & 29.9 & [23.8, 44.4] & 28.5 & [26.8, 32.6] \\
    GR00T~N1.7-DROID  & 268 (11.2) & 8.8  & 9.2  & [8.2, 19.4]  & 9.1  & [8.7, 12.7] \\
    \bottomrule
  \end{tabular}
\end{tableH}

Flags spread over many tasks: 80 of 120 for $\pi_{0.5}$ and 69 for GR00T. Flagged episodes succeed less often: 21.0\% against 29.9\% for $\pi_{0.5}$, and 5.6\% against 9.2\% for GR00T. Confirmed events fall into five kinds (\cref{tab:robolab-events}). Most false flags come from two: a shelf rack sliding on the table, and objects dropping at the start of an episode.

\begin{tableH}
  \centering
  \caption{\textbf{RoboLab events by kind.} Confirmed contact ejection events; one episode can hold several.}
  \label{tab:robolab-events}
  \small
  \begin{tabular}{lcc}
    \toprule
    Kind & $\pi_{0.5}$ & GR00T \\
    \midrule
    Object against object & 503 & 151 \\
    Shelf rack & 137 & 112 \\
    A fruit that starts overlapping its plate & 97 & 94 \\
    A robot link involved & 81 & 67 \\
    First second of the episode & 9 & 5 \\
    \midrule
    Total & 827 & 429 \\
    \bottomrule
  \end{tabular}
\end{tableH}

\paragraph{Replay.}
We replayed 110 episodes from 11 tasks and both policies. All 110 match the recording bit for bit: every body pose and joint position at every physics substep.

\paragraph{Cause.}
There are two causes. A real overlap can grow only as fast as the bodies move, so we compare the reported depth at onset with how far the two bodies moved in the step before. Of the 832 events we checked, 609 are shallower than 3\,mm and 163 fit the motion. In the other 60, from 57 episodes, the depth exceeds the motion by more than 2\,mm; in 32 of these episodes the other body is a finger. For example, a dressing bottle and a finger reach 30\,mm of overlap in one step while the bottle moves 0.1\,mm, and the bottle leaves at 2.64\,m/s. So, as on RoboDojo, GPU PhysX reports contacts far deeper than any real overlap. The second cause is a real overlap: a spoon moves 17\,mm into a plate, shows 5.2\,mm of overlap, and the engine pushes them apart.

\paragraph{Objects stuck to the gripper.}
The only miss in the human review (\cref{app:accuracy}) is a remote control stuck to the open gripper. For this class we fixed a rule before the full run. An episode is flagged if, 0.5\,s after the gripper opens, the object touches only the fingers, has no other support, and moves 10\,cm or more with them within 2\,s (\cref{tab:robolab-stick}). The rule catches the missed episode, which we used to write it. \Cref{fig:robolab-stick} shows two stuck objects. Dropping all 179 flagged episodes moves the scores from 28.1 to 29.2 and from 8.8 to 9.1.

\begin{tableH}
  \centering
  \caption{\textbf{Objects stuck to the gripper} on the 4,800 RoboLab episodes. We checked 20 flagged clips by eye, 13 of them from $\pi_{0.5}$.}
  \label{tab:robolab-stick}
  \small
  \begin{tabular}{lccc}
    \toprule
    & $\pi_{0.5}$ & GR00T & Total \\
    \midrule
    Flagged & 109 (4.5\%) & 70 (2.9\%) & 179 (3.7\%) \\
    Also flagged as contact ejection & 32 & 13 & 45 \\
    Successes among the flagged & 4 & 1 & 5 \\
    \midrule
    \multicolumn{4}{@{}l}{\emph{Checked by eye: 20 clips}} \\
    Stuck or passing through the gripper & \multicolumn{3}{c}{10, 50\% (95\% CI 30--70\%)} \\
    Normal & \multicolumn{3}{c}{10; 5 of them pack cans or boxes into a bin} \\
    \bottomrule
  \end{tabular}
\end{tableH}

\subsection{RoboDojo}
\begin{figureH}
  \centering
  \includegraphics[width=\linewidth]{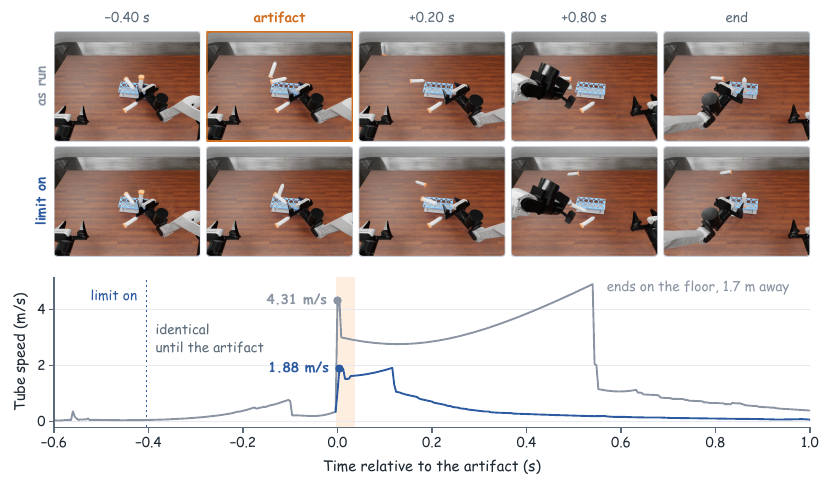}
  \caption{\textbf{RoboDojo, Insert Tubes (layout 2), XR-1.} Head camera frames around the first artifact, and the speed of the ejected tube. Top: as run. Bottom: replayed with a depenetration limit of 1\,m/s from 0.4\,s before the artifact. As run, the tube leaves the rack at 4.31\,m/s within one step and lands 1.7\,m away, off the table; the later peak of the gray curve is its fall. With the limit, it leaves at 1.88\,m/s, slows below 0.4\,m/s within 0.3\,s and stays on the table. The as-run replay matches the recording bit for bit.}
  \label{fig:robodojo}
\end{figureH}

Seven of the 40 XR-1 episodes contain contact artifacts, all failures. The first pass over seed 0 has two events on Insert Tubes (layout~2) and three on Store Tools In Toolbox (layout~1). Later passes add two on Insert Tubes (layout~2 under seed 0, at 6.30\,m/s; layout~0 under seed 2), four on Fill Pen Holder (layouts 1 and 2, seed 0) and one on Store Tools In Toolbox (layout~1, seed 1). The first detection stage confirmed 25 events in 15 episodes; the gravity stage explained 13 of them as falls. We replayed the first-pass Insert Tubes episode with the depenetration limit switched on 100 physics steps (0.4\,s) before the ejection (\cref{tab:robodojo-cap}).

The contact that starts this ejection is also false. RoboDojo runs Kit~107.3 with GPU dynamics. The tube is a 60\,g dynamic body with a convex decomposition; the rack is a static collider with a signed distance field. The tube's depenetration limit is 3\,m/s at run time. The gripper releases the tube about 120 steps before onset, and no robot link touches it after that. The tube then tips over the upper edge of the rack. At onset, three new contact points report separations of $-10.9$ to $-10.1$\,mm, with an impulse of 1.08\,N\,s at the deepest. They lie on one side of a rack wall 2 to 5\,mm thick, 7.6\,mm outside the convex hull of the tube's mesh. The tube points they imply lie on the other side of the wall. In the 14 steps before, that rack point is 6.8 to 7.6\,mm from the tube's hull, with no contact. So the engine pairs the tube with the far side of a thin wall and reports the gap plus the wall as a depth. This matches RoboTwin, but here it happens in the GPU contact generation, where the threshold of \cref{app:cause} plays no part.

\begin{tableH}
  \centering
  \caption{\textbf{RoboDojo, Insert Tubes layout~2}: exit speed of the tube, replayed with the depenetration limit switched on 0.4\,s before the ejection.}
  \label{tab:robodojo-cap}
  \small
  \begin{tabular}{lc}
    \toprule
    Depenetration limit & Exit speed (m/s) \\
    \midrule
    As run, 3\,m/s by default & 4.31 \\
    1\,m/s & 1.88 \\
    0.3\,m/s & 0.62 \\
    \bottomrule
  \end{tabular}
\end{tableH}

\subsection{RoboDojo: eight models against the real robot}
\label{app:robodojo-real}

\paragraph{Setup.}
RoboDojo publishes real robot results for models with official simulation weights. We run eight of them: InternW0, OpenWAM, $\pi_{0.5}$, InternVLA-A1, Xiaomi-Robotics-0, X-VLA, GR00T~N1.7 and $\pi_0$. We use the seven tasks that exist both in simulation and on the real robot: Classify Objects, Cover Blocks, Fill Pen Holder, Hang Mugs, Insert Tubes, Stack Bowls and Sweep Blocks. Each task runs on seed~0, with 30 layouts for Hang Mugs and 55 for the others: 360 episodes per model, 2,880 in total. We keep the official protocol, including batch sizes. Sim success is the mean over the seven tasks. Real results are the public leaderboard values.

\paragraph{Results.}
The detector flags 416 of the 2,880 episodes (14.4\%); 10 of them are successes. Insert Tubes (122) and Fill Pen Holder (97) give the most flags. We checked 100 flagged and 100 unflagged episodes by eye (\cref{app:review}). \emph{Reviewed} weights each flagged episode by the share of real artifacts for its kind, or uses its own label if checked.

\begin{tableH}
  \centering
  \caption{\textbf{RoboDojo, eight models}, sorted by real success. Sim columns in \%, over the seven shared tasks. \emph{All flags}: every flagged episode removed. \emph{Reviewed}: real artifacts removed, from the human review. Real columns: public leaderboard.}
  \label{tab:robodojo-real}
  \small
  \setlength{\tabcolsep}{4pt}
  \begin{tabular}{@{}lccccccc@{}}
    \toprule
    & & \multicolumn{4}{c}{Sim} & \multicolumn{2}{c}{Real} \\
    \cmidrule(lr){3-6}\cmidrule(l){7-8}
    Model & Flagged & Reported & All flags & Reviewed & Interval & Success & Total score \\
    \midrule
    InternW0          & 43 & 39.7 & 41.8 & 40.6 & [38.4, 50.5] & 31.7 & 38.9 \\
    OpenWAM           & 73 & 19.4 & 21.2 & 19.8 & [18.9, 40.9] & 24.4 & 37.6 \\
    $\pi_{0.5}$       & 65 & 17.1 & 17.6 & 17.0 & [16.4, 35.8] & 12.8 & 22.9 \\
    InternVLA-A1      & 43 & 2.1  & 2.2  & 2.1  & [2.1, 14.3]  & 7.2  & 12.0 \\
    Xiaomi-Robotics-0 & 51 & 5.2  & 5.3  & 5.3  & [5.2, 20.0]  & 3.9  & 7.9 \\
    X-VLA             & 56 & 15.8 & 17.5 & 16.7 & [15.8, 31.6] & 3.3  & 7.6 \\
    GR00T~N1.7        & 39 & 1.3  & 1.3  & 1.3  & [1.3, 12.3]  & 1.7  & 5.9 \\
    $\pi_0$           & 46 & 3.6  & 3.7  & 3.7  & [3.6, 17.1]  & 1.7  & 5.8 \\
    \bottomrule
  \end{tabular}
\end{tableH}

\begin{tableH}
  \centering
  \caption{\textbf{Rank correlation of sim with real}, Spearman / Kendall, eight models. Sim scores from \cref{tab:robodojo-real}: reported, all flags removed, and interval middle.}
  \label{tab:robodojo-corr}
  \small
  \begin{tabular}{lccc}
    \toprule
    Real measure & Reported & All flags & Interval middle \\
    \midrule
    Success rate      & 0.81 / 0.69 & 0.81 / 0.69 & 0.81 / 0.69 \\
    Total score       & 0.79 / 0.64 & 0.79 / 0.64 & 0.79 / 0.64 \\
    ARX~X5 score      & 0.69 / 0.57 & 0.69 / 0.57 & 0.69 / 0.57 \\
    Piper score       & 0.81 / 0.64 & 0.81 / 0.64 & 0.81 / 0.64 \\
    Piper~X score     & 0.55 / 0.36 & 0.55 / 0.36 & 0.55 / 0.36 \\
    \bottomrule
  \end{tabular}
\end{tableH}

With the reviewed scores, Spearman and Kendall against the total score stay 0.79 and 0.64. Neither correction changes the order of any pair of models, and our sim ranking equals the official one (Spearman 1.0). The pairs that disagree with real success are InternVLA-A1 against Xiaomi-Robotics-0, X-VLA and $\pi_0$, and Xiaomi-Robotics-0 against X-VLA. They disagree in the official sim scores too. So on RoboDojo, artifacts are rare in successes and do not explain the sim-to-real gap.

\paragraph{Cause.}
We took two flagged Insert Tubes events: Xiaomi-Robotics-0 on layout~23 at step 2144, and GR00T~N1.7 on layout~17 at step 2792. We saved the full state one step before onset and stepped on with changed settings. Both events behave the same; we report the first. With GPU PhysX, as run, the engine reports a contact about 13\,mm deep between the tube and the top of the rack, and the tube leaves at 2.6\,m/s. With CPU PhysX, the tube and rack do not overlap, and the tube stays below 0.6\,m/s for the next 20 steps. Modeling the rack as convex pieces instead of a triangle mesh also removes the contact. Changing or disabling the rack's signed distance field gives a bit-identical result, so it is not used for this pair. Raising the contact offset only delays the false contact. As a control, moving the rack up by 5\,mm does change the result. The false contact points lie on top of the rack, where the wall is 4.5 to 4.9\,mm thick, with the normal reversed.

\subsection{LIBERO}
\begin{figureH}
  \centering
  \includegraphics[width=\linewidth]{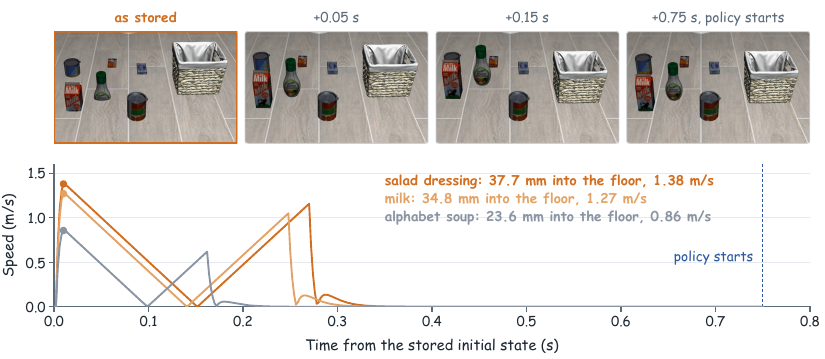}
  \caption{\textbf{LIBERO-Object task 0, stored initial state 0}, built as the evaluation builds it with official LIBERO, robosuite 1.4.0 and MuJoCo 3.2.3, and run through the 15 settle steps before the policy acts. As stored, the salad dressing, the milk and the alphabet soup sit 37.7, 34.8 and 23.6\,mm inside the floor. MuJoCo pushes them out within 10\,ms at 1.38, 1.27 and 0.86\,m/s. They hop, land within 0.3\,s and are at rest before the policy acts at 0.75\,s. Depths and peak speeds match the audit of all 2000 stored states, of which 650 start inside a support.}
  \label{fig:libero-init}
\end{figureH}
\begin{tableH}
  \centering
  \caption{\textbf{LIBERO's stored initial states}, built as the evaluation builds them and simulated for the 15 steps before the policy acts, without a policy.}
  \label{tab:libero-init}
  \small
  \begin{tabular}{lccccc}
    \toprule
    Suite & Initial states & Contact ejection & After gravity & Contact artifacts & Tasks affected \\
    \midrule
    LIBERO-Spatial & 500 & 3   & 0   & 0   & 0/10 \\
    LIBERO-Object  & 500 & 500 & 500 & 500 & 10/10 \\
    LIBERO-Goal    & 500 & 0   & 0   & 0   & 0/10 \\
    LIBERO-10      & 500 & 152 & 150 & 150 & 3/10 \\
    \bottomrule
  \end{tabular}
\end{tableH}

\paragraph{Initial states.}
In the floor scene of LIBERO-Object and the living room table of LIBERO-10 tasks 0, 1 and 7, the stored initial states place objects inside their support (\cref{fig:libero-init,tab:libero-init}). Objects sit up to 37.7\,mm into the floor and 14.6\,mm into the table. The depth is the same in every initial state of a scene, since the states differ only in horizontal positions. Within the first 10\,ms, MuJoCo launches the objects out at up to 1.38\,m/s off the floor and 1.83\,m/s off the table, with no robot contact. In the 130 policy episodes that start from these states, the policy succeeds in 96.2\%, against 94.8\% in the other 270.

\paragraph{Episodes with the policy.}
Two of the 400 episodes contain a contact artifact while the policy acts. In LIBERO-10 task 9, which puts a mug into the microwave, MuJoCo reports a contact 242.8\,mm deep with 412\,N between the mug and a plate of the microwave, while its own distance query gives 1.4\,mm. The mug leaves the gripper at 1.57\,m/s and the episode fails. In LIBERO-Object task 4, a finger touches the ketchup again during the release, and the ketchup's speed rises from 0.25 to 0.62\,m/s for one step; the episode succeeds.

\subsection{RoboCasa}

Five of the 300 XR-1 episodes contain a contact artifact, all failures. On the ten placement tasks: bread against its basket at 0.61\,m/s on Arrange Bread Basket, an object leaving a drawer at 1.19\,m/s on Pick Place Drawer To Counter, and a bowl on the floor at 0.53\,m/s on Stack Bowls Cabinet. On the other 40 tasks, with doors, drawers, appliances and long-horizon composites: a baguette against a toaster oven door at 0.61\,m/s on Heat Kebab Sandwich, and meat on the floor, pushed by the mobile base, at up to 0.84\,m/s on Pack Identical Lunches. The first detection stage flags 22 episodes; the gravity stage explains 17 of them as falls.

\section{Episode Accounting}
\label{app:manifest}

\Cref{tab:counts} gives the counts behind \cref{tab:scores}: failures with an artifact, failures that pass when replayed with the limit, and successes with an artifact that fail. It also lists the tasks outside the five. \Cref{tab:scores} leaves them out because their audit is incomplete. HyVLA's successes on its six other tasks were not checked for artifacts, and HyVLA was not run on the remaining 39 tasks. The closed-loop runs cover the five tasks and, for LingBot-VA, the six others (\cref{tab:closed-loop-depen}).

\begin{tableH}
  \centering
  \caption{\textbf{Counts behind \cref{tab:scores}.} \emph{Art.}: failures with an artifact, of all failures. \emph{Resc.}: failures that pass under the limit. \emph{Lost}: successes that fail under the limit, of all successes with an artifact. The five tasks use the short limit on the ejected body (\cref{tab:narrow}). All other entries use the limit on every body until the end. $^\dagger$Both passes of every task (\cref{tab:runs}). Of the 40 tasks, the first pass was replayed: 4 failures and 3 successes with an artifact. The three Put Object Cabinet replays of the second pass diverge from the recording before the switch and are not counted.}
  \label{tab:counts}
  \small
  \begin{tabular}{@{}lcccccc@{}}
    \toprule
    & \multicolumn{3}{c}{LingBot-VA} & \multicolumn{3}{c}{HyVLA} \\
    \cmidrule(lr){2-4}\cmidrule(l){5-7}
    Task & Art. & Resc. & Lost & Art. & Resc. & Lost \\
    \midrule
    Place Can Basket    & 21/23 & 14 & 8/16 & 1/4  & 1  & 2/3 \\
    Stack Bowls Three   & 12/12 & 2  & --   & 2/2  & 0  & 1/1 \\
    Place Object Basket & 3/5   & 2  & 2/3  & 0/1  & -- & -- \\
    Place Bread Basket  & 1/1   & 0  & --   & 0/0  & -- & -- \\
    Stack Bowls Two     & 3/3   & 3  & --   & 3/3  & 3  & -- \\
    \midrule
    Five tasks          & 40/44 & 21 & 10/19 & 6/10 & 4  & 3/4 \\
    Six other tasks     & 1/24  & 1  & --   & 0/38 & -- & -- \\
    Other 45 tasks$^\dagger$ & 12/146 & 1 & 1/9 & --   & -- & -- \\
    \midrule
    \multicolumn{7}{@{}l}{\emph{Rates outside the five tasks: as run / replay / exclusion}} \\
    Six other tasks     & \multicolumn{3}{c}{76.0 / 77.0 / 76.8} & \multicolumn{3}{c}{68.3 / 68.3 / --} \\
    Other 45 tasks$^\dagger$ & \multicolumn{3}{c}{91.9 / -- / --} & \multicolumn{3}{c}{-- / -- / --} \\
    \bottomrule
  \end{tabular}
\end{tableH}

\Cref{tab:manifest} lists every scored episode with a contact artifact. Each entry gives the seed, the recorded outcome, and the outcome of a replay with the depenetration limit of 1\,m/s switched on 0.1\,s before the first artifact. Summing $r_e - y_e$ over these rows gives the rate with the limit on every body until the end: 91.9\% for LingBot-VA and 93.0\% for HyVLA on the five tasks. \Cref{tab:narrow} gives the short form. Place Can Basket ran three repetitions of 30 seeds (\cref{app:runs}), so a seed can appear more than once.

\begin{tableH}
  \centering
  \caption{\textbf{Every scored episode with a contact artifact} and its replay. S: success, F: failure. The arrow leads from the recorded outcome $y_e$ to the replayed outcome $r_e$.}
  \label{tab:manifest}
  \footnotesize
  \begin{tabular}{@{}lr>{\raggedright\arraybackslash}p{0.60\linewidth}@{}}
    \toprule
    Task & Episodes & Seed, recorded $\to$ replayed \\
    \midrule
    \multicolumn{3}{@{}l}{\emph{LingBot-VA}} \\
    Place Can Basket (rep.\,1) & 13 & 100000\,F$\to$S, 100002\,S$\to$S, 100004\,S$\to$F, 100006\,F$\to$S, 100007\,F$\to$S, 100008\,F$\to$S, 100014\,S$\to$S, 100017\,S$\to$F, 100021\,F$\to$S, 100022\,S$\to$S, 100024\,S$\to$F, 100029\,F$\to$S, 100031\,F$\to$F \\
    Place Can Basket (rep.\,2) & 11 & 100004\,S$\to$S, 100011\,F$\to$S, 100012\,F$\to$S, 100017\,S$\to$F, 100019\,F$\to$S, 100021\,S$\to$S, 100022\,F$\to$S, 100025\,S$\to$F, 100026\,S$\to$S, 100027\,S$\to$F, 100029\,F$\to$F \\
    Place Can Basket (rep.\,3) & 13 & 100000\,S$\to$S, 100003\,F$\to$S, 100009\,F$\to$S, 100016\,F$\to$F, 100019\,F$\to$S, 100021\,F$\to$S, 100022\,F$\to$S, 100023\,F$\to$S, 100027\,F$\to$S, 100028\,F$\to$S, 100029\,S$\to$F, 100030\,S$\to$S, 100032\,S$\to$S \\
    Stack Bowls Three & 12 & 100002\,F$\to$S, 100009\,F$\to$F, 100023\,F$\to$F, 100028\,F$\to$F, 100047\,F$\to$S, 100067\,F$\to$S, 100071\,F$\to$S, 100073\,F$\to$F, 100075\,F$\to$S, 100077\,F$\to$S, 100078\,F$\to$S, 100089\,F$\to$F \\
    Place Object Basket & 6 & 100013\,F$\to$S, 100015\,S$\to$S; pass 2: 100001\,F$\to$S, 100009\,S$\to$F, 100014\,S$\to$F, 100016\,F$\to$S \\
    Place Bread Basket & 1 & 100003\,F$\to$S \\
    Stack Bowls Two & 3 & 100011\,F$\to$S; pass 2: 100014\,F$\to$S, 100027\,F$\to$S \\
    Dump Bin Bigbin & 1 & pass 2: 100010\,F$\to$F \\
    Put Object Cabinet & 3 & pass 2: 100110\,S$\to$S, 100129\,S$\to$S, 100210\,F$\to$S; replays diverge before the switch, not counted \\
    Place Fan & 1 & 100015\,F$\to$F \\
    Hanging Mug & 1 & 100003\,F$\to$F \\
    Stamp Seal & 1 & 100026\,F$\to$S \\
    Move Stapler Pad & 1 & 100018\,F$\to$F \\
    Place Burger Fries & 2 & 100012\,S$\to$S, 100015\,S$\to$S \\
    Place Dual Shoes & 1 & 100015\,S$\to$F \\
    \midrule
    \multicolumn{3}{@{}l}{\emph{HyVLA}} \\
    Place Can Basket & 4 & 100004\,S$\to$F, 100021\,S$\to$F, 100022\,F$\to$S, 100023\,S$\to$S \\
    Stack Bowls Three & 3 & 100016\,S$\to$S, 100022\,F$\to$S, 100023\,F$\to$F \\
    Stack Bowls Two & 3 & 100020\,F$\to$S, 100021\,F$\to$S, 100026\,F$\to$S \\
    \bottomrule
  \end{tabular}
\end{tableH}

\section{Human Verification Protocol}
\label{app:labels}

For Place Can Basket, we rendered 60 clips from recorded episodes: 40 centered on a detected event and 20 hard negatives. The hard negatives are the fastest object motion in episodes without an event, and detector flags that were never confirmed. Five robotics engineers saw the clips in random order, without their origin. They rated each clip alone, then settled disagreements by discussion; we report the agreed labels. For each clip they answered two questions: does any object move in a physically implausible way, and was the robot acting sensibly for the task before that moment? The robot was on track in 38 of the 40 events. For six more tasks, we showed 60 full evaluation videos in random order, without seed or outcome. The questions were: is there implausible motion anywhere in the episode, ignoring objects that fall after release; does the robot act sensibly; and is the task completed? Their completion labels matched the benchmark in 59 of 60 videos.

\section{Detector Accuracy}
\label{app:accuracy}

\subsection{Human review on RoboLab and RoboDojo}
\label{app:review}

On each platform we drew two sets of 100 episodes. Set~P holds flagged episodes, drawn by kind. On RoboLab the kinds are the event kinds of \cref{tab:robolab-events}. On RoboDojo there are three kinds, set by the contact at onset: shallower than 3\,mm, a depth that grows more than 2\,mm beyond what the two bodies moved, and other deep contacts. Set~R holds unflagged episodes. On RoboLab, per policy, we took 50 random tasks and one random unflagged episode from each. On RoboDojo we took 100 random unflagged episodes. One of us watched all 400 episodes. For set~P, the question was whether the flagged motion is physically implausible. For set~R, it was whether the episode has any implausible motion that the detector missed. Precision is weighted by how many flagged episodes each kind stands for. Brackets are 95\% intervals from a uniform prior per kind.

\begin{tableH}
  \centering
  \caption{\textbf{Human review, 400 episodes.} Set~P: real artifacts among checked flagged episodes, per kind. Set~R: missed artifacts among unflagged episodes.}
  \label{tab:review}
  \small
  \begin{tabular}{@{}lccc@{}}
    \toprule
    & \multicolumn{2}{c}{RoboLab} & RoboDojo \\
    \cmidrule(lr){2-3}\cmidrule(l){4-4}
    Set~P, real of checked & $\pi_{0.5}$ & GR00T & Eight models \\
    \midrule
    Object against object & 8/28 & 3/8 & -- \\
    Fruit overlapping its plate & 0/11 & 1/13 & -- \\
    A robot link involved & 3/11 & 9/10 & -- \\
    Shelf rack & 0/9 & 0/6 & -- \\
    First second & 0/2 & 1/2 & -- \\
    Depth grows faster than the motion & -- & -- & 15/18 \\
    Other deep contact & -- & -- & 9/20 \\
    Shallower than 3\,mm & -- & -- & 8/62 \\
    \midrule
    Precision, weighted & 20.2 [14.4, 34.9] & 35.9 [26.9, 50.9] & 29.9 [23.4, 37.9] \\
    \midrule
    Set~R, missed of checked & 1/50 & 0/50 & 0/100 \\
    Miss rate, 95\% upper & 10.3 & 7.0 & 2.9 \\
    \bottomrule
  \end{tabular}
\end{tableH}

Over both policies, 26\% of flagged RoboLab episodes are real artifacts; on RoboDojo, 30\%. The one miss is an object stuck to the gripper (\cref{tab:robolab-stick}). The detector picks the episodes for a person to check.

\paragraph{Effect on the score.}
We count each flagged kind at its measured share of real artifacts. We then add the estimated misses at the success rate of unflagged episodes. This gives the reviewed columns of \cref{tab:robolab}: 28.5 [26.8, 32.6] and 9.1 [8.7, 12.7]. With the miss rate at its 95\% upper bound, the intervals are [24.9, 37.2] and [8.1, 18.4]. Removing every flag instead of the reviewed share raises the score by at most 1.4 points on RoboLab and RoboDojo (\cref{tab:robolab,tab:robodojo-real}).

\subsection{Threshold sensitivity}
\label{app:sens}

We reran the contact ejection stage offline on all 4,800 RoboLab recordings. The frozen thresholds reproduce the online events exactly. We then scaled one threshold at a time by 0.5, 0.75, 1.5 and 2, and all six together by 0.75 and 1.5. This gives 26 settings (\cref{tab:sens}). The flag rate changes a lot, but the exclusion score moves by at most 1.6 points for $\pi_{0.5}$ and 0.3 for GR00T.

\begin{tableH}
  \centering
  \caption{\textbf{Threshold sensitivity on RoboLab}, all flags. Each row gives the range over the four scales of one threshold; the last row covers all 26 settings. Flag rate and exclusion score in \%.}
  \label{tab:sens}
  \small
  \begin{tabular}{lcccc}
    \toprule
    & \multicolumn{2}{c}{$\pi_{0.5}$} & \multicolumn{2}{c}{GR00T} \\
    \cmidrule(lr){2-3}\cmidrule(l){4-5}
    Threshold, $\times$0.5 to $\times$2 & Flag rate & Exclusion & Flag rate & Exclusion \\
    \midrule
    Frozen & 20.7 & 29.9 & 11.2 & 9.2 \\
    Minimum speed            & 9.7--23.2  & 29.0--30.6 & 5.5--12.5  & 9.1--9.3 \\
    Minimum velocity change  & 10.0--24.8 & 29.2--30.5 & 4.5--13.2  & 9.1--9.4 \\
    Minimum displacement     & 14.8--21.7 & 29.5--30.0 & 6.7--12.0  & 9.1--9.3 \\
    Free flight              & 16.1--21.9 & 29.7--30.4 & 7.7--11.3  & 9.2--9.3 \\
    Violent contact, velocity change & 20.6--21.0 & 29.9--30.0 & 11.1--11.4 & 9.2--9.3 \\
    Violent contact, speed   & 20.6--20.8 & 29.9--30.0 & 11.1--11.2 & 9.2 \\
    All six, $\times$0.75 and $\times$1.5 & 9.6--26.7 & 29.0--30.6 & 4.9--13.9 & 9.2--9.3 \\
    \midrule
    All 26 settings & 9.6--26.7 & 29.0--30.6 & 4.5--13.9 & 9.1--9.4 \\
    \bottomrule
  \end{tabular}
\end{tableH}

\subsection{Baselines}
\label{app:baselines}

We compare with three simple rules on the same 4,800 RoboLab episodes. Each rule reads the same bodies as our detector and flags an episode when one value passes a threshold. The values are the top speed, the largest velocity jump in one substep, and the largest penetration depth. Labels come from the 200 RoboLab episodes of \cref{app:review}, each weighted by the share of the population it stands for. \Cref{tab:baselines} sets every rule to the flag rate of SimuGuard, 15.9\%.

\begin{tableH}
  \centering
  \caption{\textbf{Baselines on RoboLab} at the same flag rate, 15.9\%. Precision, recall and F1 in \%, weighted by the population.}
  \label{tab:baselines}
  \small
  \begin{tabular}{lccc}
    \toprule
    Method & Precision & Recall & F1 \\
    \midrule
    SimuGuard, frozen thresholds & \textbf{26} & \textbf{84} & \textbf{39} \\
    Top speed & 20 & 65 & 30 \\
    Largest velocity jump per substep & 19 & 62 & 29 \\
    Largest penetration depth & 13 & 34 & 19 \\
    \bottomrule
  \end{tabular}
\end{tableH}

Common fixed thresholds flag far more episodes. A top speed above 1\,m/s flags 43.1\% of episodes, with precision 10. Tuned on the labels, the best speed rule reaches F1 41 and the best penetration rule 22.

\subsection{Oracle with the repaired engine on RoboTwin}
\label{app:oracle}

On the five RoboTwin tasks we replayed all 260 LingBot-VA episodes open loop with their recorded actions. The replay uses SAPIEN with the PhysX~5.4.0 threshold (\cref{app:cause}). In this repaired engine, a missed artifact that changed the outcome would show up as an unflagged failure that now succeeds.

\begin{tableH}
  \centering
  \caption{\textbf{Repaired-engine oracle}, RoboTwin, five tasks, LingBot-VA, open-loop replay.}
  \label{tab:oracle}
  \small
  \begin{tabular}{lcccc}
    \toprule
    Episodes & Count & Successes as run & Failure $\to$ success & Success $\to$ failure \\
    \midrule
    Unflagged & 201 & 197 & 3/4 & 27/197 \\
    Flagged & 59 & 19 & 34/40 & 9/19 \\
    \bottomrule
  \end{tabular}
\end{tableH}

At most 3 unflagged episodes, 1.5\%, are missed artifacts that changed the outcome. Of the 40 flagged failures, 34 succeed. The new threshold also changes ordinary grasp contacts, so every replay drifts from the recording at some point. So 27 unflagged successes fail, 20 of them on Place Can Basket.

\end{document}